\documentclass[conference]{IEEEtran}
\usepackage{times}

\usepackage[numbers]{natbib}
\usepackage{multicol}
\usepackage[bookmarks=true]{hyperref}

\usepackage{amsmath}
\usepackage{graphicx}
\usepackage{algorithm}
\usepackage{algpseudocode}
\usepackage{multirow}
\usepackage{booktabs}
\usepackage[dvipsnames]{xcolor}
\usepackage{hyperref}
\usepackage{bbm}
\hypersetup{
    colorlinks=true,
    linkcolor=orange,
    filecolor=orange,      
    urlcolor=orange,
    citecolor=orange,
}
\usepackage[capitalise, nameinlink]{cleveref}
\usepackage{url}
\usepackage[font=small]{caption}
\usepackage{tabularx}
\usepackage{wrapfig}
\usepackage{booktabs}
\usepackage{multirow}
\usepackage{listings}
\usepackage{xcolor}
\usepackage{inconsolata}

\usepackage{amsmath,amsfonts,bm}

\def\eqref#1{equation~\ref{#1}}

\def\1{\bm{1}}

\DeclareMathAlphabet{\mathsfit}{\encodingdefault}{\sfdefault}{m}{sl}
\SetMathAlphabet{\mathsfit}{bold}{\encodingdefault}{\sfdefault}{bx}{n}

\newcommand{\E}{\mathbb{E}}

\newcommand{\softmax}{\mathrm{softmax}}

\DeclareMathOperator*{\argmax}{argmax}

\newcommand{\arl}{a^\text{RL}}
\newcommand{\ail}{a^\text{IL}}

\definecolor{codegreen}{rgb}{0,0.6,0}
\definecolor{codegray}{rgb}{0.5,0.5,0.5}
\definecolor{codepurple}{rgb}{0.58,0,0.82}
\definecolor{backcolour}{rgb}{0.95,0.95,0.95}

\lstdefinestyle{mystyle}{
    backgroundcolor=\color{backcolour},   
    commentstyle=\color{codegreen},
    keywordstyle=\color{magenta},
    numberstyle=\tiny\color{codegray},
    stringstyle=\color{codepurple},
    basicstyle=\ttfamily\footnotesize,
    columns=fullflexible,
    xleftmargin=0.5cm,
    frame=tlbr,
    framesep=8pt,
    framerule=0pt,
    breakatwhitespace=false,         
    breaklines=true,                 
    captionpos=b,                    
    keepspaces=true,                 
    numbers=left,                    
    numbersep=5pt,                  
    showspaces=false,                
    showstringspaces=false,
    showtabs=false,                  
    tabsize=2
}

\begin{document}

\title{Imitation Bootstrapped Reinforcement Learning}

\author{
Hengyuan Hu\\Stanford University
\and Suvir Mirchandani\\Stanford Univeristy
\and Dorsa Sadigh\\Stanford University
}

\maketitle

\begin{abstract}
Despite the considerable potential of reinforcement learning (RL), robotic control tasks predominantly rely on imitation learning (IL) due to its better sample efficiency.
However, it is costly to collect comprehensive expert demonstrations that enable IL to generalize to all possible scenarios, and any distribution shift would require recollecting data for finetuning.
Therefore, RL is appealing if it can build upon IL as an efficient autonomous self-improvement procedure.
We propose \emph{imitation bootstrapped reinforcement learning} (IBRL), a novel framework for sample-efficient RL with demonstrations that first trains an IL policy on the provided demonstrations and then uses it to propose alternative actions for both online exploration and bootstrapping target values. 
Compared to prior works that oversample the demonstrations or regularize RL with an additional imitation loss, IBRL is able to utilize high quality actions from IL policies since the beginning of training, which greatly accelerates exploration and training efficiency.
We evaluate IBRL on 6 simulation and 3 real-world tasks spanning various difficulty levels. IBRL significantly outperforms prior methods and the improvement is particularly more prominent in harder tasks.

\end{abstract}

\IEEEpeerreviewmaketitle

\section{Introduction}

Despite achieving remarkable performance in many simulation domains~\citep{silver2017mastering, vinyals2019grandmaster, fair2022human}, reinforcement learning (RL) has not been widely used in solving robotics and low level continuous control problems, especially in the real world.
The main challenges of applying RL to continuous control problems are exploration and sample efficiency.
In these settings, reward signals are often sparse by nature, and unlike learning in games where the sparse reward is often achievable within a fixed horizon, a randomly initialized neural policy may never finish a task, resulting in no signals for learning. 
Besides the hard exploration problem, RL often needs a large number of samples to converge, which hinders its adoption in the real world where massive parallel simulation is not available.

As a result, most learning-based robotics systems rely on imitation learning (IL)~\citep{brohan2023rt-1} or offline RL~\citep{Kumar2022PreTrainingFR} with strong assumptions such as access to large specialized datasets. 
However, those methods come with their own challenges. Expert demonstrations are often expensive to collect and require access to expert operators and domain knowledge~\citep{mandlekar2021what}.
In addition, policies learned from static datasets suffer from distribution shifts when deployed in slightly different environments. 
Given these challenges, online RL algorithms -- when carefully integrated with IL -- can still play a valuable role in efficiently learning robot policies. 
An ideal RL algorithm for real world robotics applications should be able to benefit from human demonstrations and strong IL methods for sample-efficient learning. Moreover, it should go far beyond these IL techniques via self-improvement to reach higher performance or to address distribution shift.

The most straightforward way to use demonstration data in RL is to initialize the RL replay buffer with demonstrations and oversample those demonstrations during training~\citep{vecerik2017leveraging}.
This approach does not leverage the fact that IL policies trained on the demonstrations can indeed provide more useful information -- they can output actions that 
may not be good enough to solve unseen scenarios, but can still provide some ``lower bound" on the action quality when the initial RL actions are highly suboptimal.
Another common approach is to pretrain the RL policy with human data and then fine-tune it with RL while applying additional regularization~\cite{haldar2022watch-rot} to ensure that the knowledge from demonstrations does not get washed out quickly by the randomly initialized critics.
This approach requires balancing the primary RL loss and the secondary IL regularization loss to achieve maximum performance, which may require hyper-parameter tuning that is infeasible in the real world.
Additionally, this necessitates using the same architecture to fit IL and RL data, which is undesirable in complex tasks as RL and IL may require very different architectures.

\begin{figure*}[t]
\centering
\includegraphics[width=0.75\linewidth]{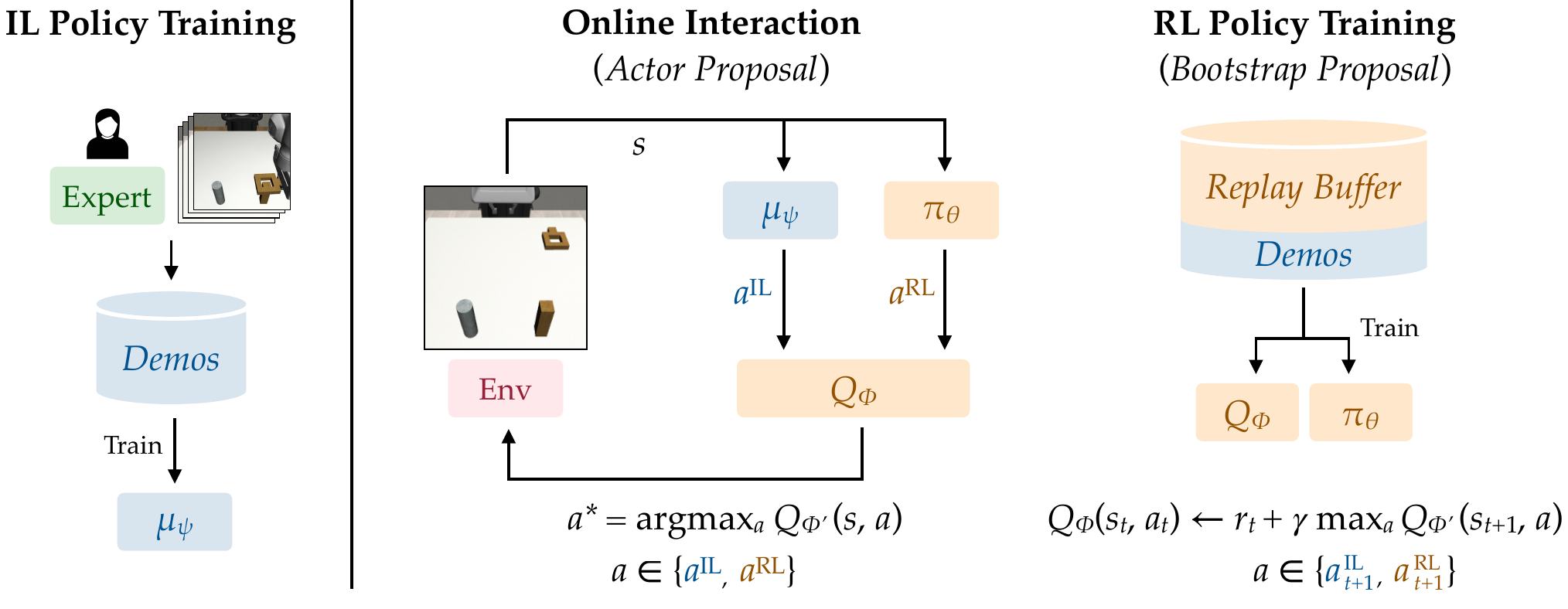}
\vspace{-1mm}
\caption{\small 
\textbf{Imitation-Bootstrapped Reinforcement Learning (IBRL).} 
IBRL first trains an imitation learning policy and then uses it to propose additional actions for RL during both online interaction phase (actor proposal) and training phase (bootstrap proposal).
We use the moving average of the online Q-function, i.e. the target Q-function $Q_{\phi'}$, to decide which action to take. 
}
\label{fig:algo}
\vspace{-5mm}
\end{figure*}

We propose \emph{imitation bootstrapped reinforcement learning} (IBRL), a method to effectively combine IL and RL for sample-efficient reinforcement learning. 
IBRL first trains a separate, standalone imitation policy on the provided demonstrations with a powerful neural network that is much deeper than the ones normally used in online RL. 
Then IBRL explicitly uses this IL policy in two phases to accelerate RL training.
First, during the online interaction phase, both the IL policy and RL policy propose an action and the agent executes the action that has a higher Q-value according to the Q-function being trained by the RL.
Second, during the training phase of RL, the target for updating the Q-values again bootstraps from the better action among the ones proposed by either the RL or the IL policies. 
Similar to prior work, we also pre-fill the RL replay buffer with the demonstrations to provide learning signals before the policy collects its first online success.
\cref{fig:algo} illustrates the core idea of IBRL, and how an IL policy is explicitly integrated in the interaction and training phase of RL. 
By keeping the IL policy separate, IBRL does not need explicit regularization loss to prevent catastrophic forgetting and thus eliminate the need to search for proper hyperparameters to balance RL and IL.
It also allows the IL to utilize deeper, more powerful networks that may be hard to train in RL with sparse reward. 
By explicitly considering actions from the IL policy, IBRL improves the quality of exploration and value estimation when the RL policy is inferior. 
It may also benefit from any potential generalizations of the IL policy in states beyond the limited demonstration data.

We evaluate IBRL on 6 simulation and 3 real-world robotics tasks spanning various difficulty levels. 
All tasks use sparse 0/1 reward.
IBRL matches or outperforms strong existing methods on all tasks and the improvement is more significant in harder tasks. In particular, IBRL nearly doubles the performance over the second best method in the hardest simulation task evaluated in this paper.
In a challenging real-world deformable cloth hanging task, IBRL performs 2.4$\times$ better than the second best RL method. In fact, prior methods are unable to even surpass the BC baseline after 2 hours of real-world training on this task.


\section{Related Work}
\label{sec:related}

In this section, we review methods that address the sample efficiency of RL both with and without access to human demonstrations. We also cover a particularly relevant area of work that uses a reference policy in RL for various purposes.

\smallskip \noindent \textbf{Sample-Efficient RL.} 
A number of recent works have greatly improved sample efficiency of RL by applying various regularization techniques.
For instance, RED-Q~\citep{chen2021randomized} and Dropout-Q~\citep{hiraoka2022dropout} apply regularization to the Q-function (critics) via ensembling or dropout so that they can be trained with higher update-to-data (UTD) ratio (i.e., the number of updates for every transition collected), leading to faster convergence and thus higher sample efficiency.
These approaches are commonly used in state-based RL, where it is computationally feasible to have a large number of independent critics made of shallow fully connected layers.
For learning directly from pixel inputs, image augmentation such as random shifts \citep{yarats2022mastering} can instead boost performance and sample efficiency without the need of increasing UTD ratio and thus maintains low computational cost. 
We apply RED-Q and image augmentation in our method, IBRL, for state- and pixel-based experiments respectively to build upon these strong foundations.

\smallskip \noindent \textbf{RL with Prior Demonstrations.} 
In sparse reward settings, sample-efficient RL algorithms alone are insufficient because they are unlikely to collect any reward signal through random exploration. A common approach is to supply RL with successful prior data or human demonstrations so that it has some initial signals to learn from.
The most straightforward approach that leverages demonstrations in RL is to include the demonstrations in the replay buffer and oversample the demonstrations during training with an off-policy RL algorithm~\citep{vecerik2017leveraging}. 
Despite its simplicity, \citet{ball2023efficient} recently have shown that this approach -- Reinforcement Learning from Prior Data (RLPD) --  when combined with modern sample efficient RL techniques such as  normalization, Q-ensembling, and image augmentation, outperforms many more complex RL algorithms in continuous control domains that utilize prior data. 
Meanwhile, \citet{song2023hybrid} provide theoretical analysis of a similar idea (Hybrid RL) and show that it is both effective and sample efficient. 

Another commonly used approach is to pretrain the RL policy with demonstration data and then fine-tune it with online RL~\cite{hester2018deep, rajeswaran2018learning, nair2018overcoming}. 
During RL fine-tuning, regularization is required to avoid catastrophic forgetting caused by undesirable learning signals from randomly initialized critics. Approaches such as Regularized Optimal Transport (ROT)~\cite{haldar2022watch-rot} extend this idea to visual observations and integrate an optimal transport reward as well as adaptive weighting over the regularization loss. This regularized fine-tuning approach achieves strong results in simulation and real-world robotic tasks.

Apart from model-free RL, model-based RL is also well-positioned to use prior data. 
MoDem~\cite{hansen2023modem} is a model-based planning/RL method that uses demonstrations to pretrain the policy via behavioral cloning and then pretrains the world model and critic using demonstrations as well as rollouts from the pretrained BC policy.
It then uses TD-MPC, a model predictive control (MPC) style planning algorithm augmented by Q-functions, to generate action for online inference and update the Q-functions with temporal difference (TD) learning.
MoDem compares favorably to a number of prior RL with demonstrations algorithms \citep{rajeswaran2018learning, hafner2021mastering, seo2022masked, zhan2022learning}.

Compared to the three families of methods listed above, the uniqueness of our method, IBRL, stems from the use of a powerful, standalone IL policy that provides alternative high quality actions during both inference and training. 
In IBRL, the IL policy is directly integrated into the learning algorithm so that we no longer need to arbitrarily oversample demonstrations to overweight those learning signals.
Additionally, because the IL policy is separate and will not be modified by RL gradients, IBRL eliminates the need for a carefully scheduled regularization loss that prevents the policy from forgetting. 
This further allows for the RL and IL policies to use their own most suitable network architectures and loss formulations. 
Lastly, compared to the model-based approaches, IBRL achieves strong performance while incurring significantly lower computational cost, which makes it more suitable for high frequency control in the real world.
As we show later in \cref{sec:experiments}, IBRL achieves superior performance over these alternative techniques.


\smallskip \noindent \textbf{Reference Policy in RL.} 
Similar to IBRL, many prior works in RL and search have utilized a standalone policy (reference policy) trained on human demonstrations that is separate from the policy being trained online for various purposes.
In human-AI coordination, reference policies trained from human data~\cite{bakhtin2023mastering, pikl} or induced from large language models~\cite{hu2023language} are used to regularize RL policy updates to stay close to human-like equilibria.
In robot learning, prior works have used reference policies during online interaction to assist exploration. 
EfficientImitate~\cite{efficientimitate} uses a fixed BC policy to propose action candidates for Monte Carlo Tree Search (MTCS) alongside actions from the policy being trained during online exploration.
PEX (Policy Expansion)~\cite{zhang2023policy} samples actions from a mixture of online RL policy and a reference offline RL policy during online exploration of RL.
In comparison, IBRL uses the IL reference policy in both exploration and training stages and we find it crucial to have both stages to achieve maximum sample efficiency and final performance.
In addition, none of these prior works have been evaluated in real-world robot tasks, and PEX is only evaluated with low dimensional state inputs. We evaluate IBRL in real world robot tasks as well as simulations with both image and state inputs.

%


\section{Background}

We consider a standard Markov decision process (MDP) consisting of state space $s \in \mathcal{S}$, continuous action space $\mathcal{A} = [-1, 1]^{d}$, deterministic state transition function $\mathcal{T}: \mathcal{S} \times \mathcal{A} \rightarrow \mathcal{S}$, sparse reward function $\mathcal{R}: \mathcal{S} \times \mathcal{A} \rightarrow \{0, 1\}$ that returns $1$ when the task is completed and $0$ otherwise, and discount factor $\gamma$.

\smallskip \noindent \textbf{Reinforcement Learning.} 
IBRL builds on off-policy RL methods as they can easily consume demonstration data generated by humans. 
Deep RL methods for continuous action spaces jointly learn a policy (actor) $\pi_\theta$ and one or multiple value functions (critic) $Q_\phi$ parameterized by neural networks $\theta$ and $\phi$ respectively. The value functions $Q_\phi$ are trained to minimize TD-error $L(\phi) = [r_t + \gamma Q_{\phi'}(s_{t+1}, \pi_{\theta'}(s_{t+1})) - Q_{\phi}(s_t, a_t)]^2$ while the policy is trained to output actions with high Q-values with $L(\theta) = -Q_{\phi}(s, \pi_\theta(s))$. $\pi_{\theta'}$ and $Q_{\phi'}$ are target networks whose parameters $\theta'$, $\phi'$ are exponential moving averages of $\theta$, $\phi$ respectively.


\smallskip \noindent \textbf{Imitation Learning.}
We assume access to a dataset $\mathcal{D}$ of demonstrations collected by expert human operators. Each trajectory $\xi \in \mathcal{D}$ consists of a sequence of transitions $\{(s_0, a_0), \dots, (s_T, a_T)\}$.
The most common IL method is behavior cloning (BC) which trains a parameterized policy $\mu_\psi$ to minimize the negative log-likelihood of data, i.e., $L(\psi) = -\E_{(s,a)\sim\mathcal{D}}[\log \mu_\psi(a | s)]$.
In this work, we assume $\mu_\psi$ follows an isotropic Gaussian as its action distribution for simplicity. 
We note that our framework can easily accommodate more powerful IL methods such as BC-RNN with a Gaussian mixture model~\citep{mandlekar2021what}.
With the isotropic assumption, the BC training objective for the policy can be formulated as the following squared loss: $L(\psi) = \E_{(s,a)\sim\mathcal{D}}\left\lVert\mu_\psi(s) - a \right\rVert _2^2$.

\section{Imitation Bootstrapped RL}






\subsection{Core Algorithm}

The core idea of IBRL is to first train an IL policy $\mu_{\psi}$ using expert demonstrations and then leverage this standalone reference IL policy in two phases in RL: 1) to help exploration during the online interaction, and 2) to help with target value estimation in TD learning (as shown in~\cref{fig:algo}). 
We refer to the first phase as \emph{actor proposal} and the second phase as \emph{bootstrap proposal}.

We focus our discussion on off-policy RL methods since they often have higher sample efficiency by effectively reusing past experiences as well as human demonstrations. 
Most popular off-policy RL methods for continuous control, such as Soft Actor-Critic (SAC)~\citep{haarnoja2018sac} or Twin Delayed DDPG (TD3)~\citep{fujimoto2018addressing} involve training Q-networks to evaluate the action quality and training a separate policy network to generate actions with high Q-values.
In IBRL, \emph{actor proposal} generates additional actions alongside the RL policy to assist with exploration while \emph{bootstrap proposal} accelerates Q-network training.

\smallskip \noindent \textbf{Online Interaction: \emph{Actor Proposal.}} In sparse reward robotics tasks, such as picking up a block and receiving reward only when the block is picked up, randomly initialized Q-networks and policy networks may hardly obtain any successes even after a long period of interaction, resulting in no signal for learning.
IBRL helps mitigate the exploration challenge by using a standalone IL policy $\mu_{\psi}$ trained on human demonstrations $\mathcal{D}$. 
IBRL uses this reference IL policy to propose an alternative action $\ail \sim \mu_\psi(s)$ in addition to the action $\arl \sim \pi_\theta(s)$ proposed by the RL policy at each online interaction step. Then, IBRL queries the target Q-network $Q_{\phi'}$ and selects the action with higher Q-value between the two candidates. 
That is, during online interaction, IBRL takes an action that provides the higher Q-value between the one proposed by the imitation policy $\mu_\psi$ and the one proposed by the RL policy $\pi_\theta$ that is being trained:
\begin{equation}
a^* = \argmax_{a\in \{a^{\text{IL}}, a^{\text{RL}}\}} Q_{\phi'} (s,a).
\label{eq:act}
\end{equation}
This is the \emph{actor proposal} phase of IBRL (\cref{fig:algo} middle).

\smallskip \noindent \textbf{RL Training: \emph{Bootstrap Proposal}.} 
Similarly, when computing the training targets for the Q-networks, instead of bootstrapping from $Q_{\phi'}(s_{t+1}, \pi_{\theta'}(s_{t+1}))$, we can bootstrap from the higher value between $Q_{\phi'} (s_{t+1}, \ail_{t+1})$ and $Q_{\phi'} (s_{t+1}, \arl_{t+1})$ where $\ail_{t+1}$ is sampled from the imitation policy while $\arl_{t+1}$ is sampled from the target actor $\pi_{\theta'}$:
\begin{equation}
    Q_\phi (s_t, a_t) \leftarrow r_t + \gamma \max_{a' \in \{\ail_{t+1}, \arl_{t+1}\}} Q_{\phi'} (s_{t+1},a').
\label{eq:train}
\end{equation}
This essentially assumes that the future rollout will be carried out by a policy that always picks the action between $\{\ail, \arl\}$ with the higher Q-value for every time step, which is precisely the greedy version of the exploration policy in IBRL. We refer to this phase of IBRL as \emph{bootstrap proposal} (\cref{fig:algo} right). 

In summary, IBRL replaces the policy $\pi_\theta$ in vanilla RL algorithms with a hybrid policy $\argmax_{a\in \{a^{\text{IL}}, a^{\text{RL}}\}} Q_{\phi'} (s,a)$ in both inference and training.
The idea of IBRL can be combined with any actor-critic style off-policy RL algorithm such as TD3 or SAC. 
In this paper, we use TD3 as our RL backbone because it has demonstrated strong performance and high sample efficiency in challenging RL from image settings~\cite{yarats2022mastering}.
Similar to prior works, we initialize the replay buffer with demonstrations but do not oversample those demonstrations.
We provide detailed pseudocode of IBRL with TD3 backbone in Appendix.

\smallskip \noindent \textbf{Soft IBRL Variant.} 
The discussion so far focuses on a greedy instantiation of IBRL that always selects the action with the higher Q-value. Although we find that this instantiation works well in practice -- especially in the realistic settings where the model processes raw pixels with deep image encoders -- it is worth noting that, in theory, this method may get stuck in a local optimum. 

Consider a tabular setting where the update of one $Q(s,a)$ does not lead to changes in other Q-values; then the Q-value of the optimal action $Q(s, a^*)$ will never be updated if its initial value is smaller than $Q(s, \ail)$, leading to a suboptimal solution. This problem, however, can be easily circumvented by using a \emph{soft} variant of IBRL that samples actions according to a Boltzmann distribution over Q-values instead of taking the $\argmax$, i.e., changing \cref{eq:act} of actor proposal to 
\begin{equation}
a^{*} \sim p_Q(a)
\end{equation}
and changing \cref{eq:train} of bootstrap proposal to 
\begin{equation}
Q_\phi (s_t, a_t) \leftarrow r_t + \gamma Q_{\phi'} (s_{t+1},a'), ~a' \sim p_Q(a_{t+1}), 
\end{equation}
where $p_Q(a) \propto \exp (\beta Q(s, a))$ for $a \in \{\ail, \arl\}$ with $\beta \geq 0$ being the inverse of the temperature that controls the sharpness of the distribution. 

Essentially, soft IBRL replaces the $\argmax$ operation  with a $\softmax$ to avoid the possibility of masking out optimal actions. In practice, we find this soft version works better than the normal IBRL in the state-based settings.
However, this is not essential in the more realistic pixel-based settings, possibly because with deep image encoders, changing the Q-value for certain observation-action pairs will likely cause changes to the Q-values of many other correlated inputs, which brings sufficient stochasticity to the learning process and thus mitigates the masking effect. We demonstrate the effectiveness of soft IBRL in state-based experiments in \cref{sec:exp-robomimic} while using the normal $\argmax$ version for all pixel-based experiments due to its simplicity and the fact that it does not require additional hyperparameter tuning.

\subsection{Benefits of IBRL}
\label{sec:benefits}

When using RL with access to prior demonstrations, recent work has shown that straightforward approaches such as oversampling the demonstrations as in RLPD or Hybrid RL~\cite{ball2023efficient, song2023hybrid} and BC pretraining followed by RL with BC regularization on the policy in approaches such as ROT~\cite{haldar2022watch-rot} are powerful techniques that are commonly used in real world robotics settings due to their simplicity, performance, and robustness.
In this section, we will discuss how IBRL's way of integrating IL with RL introduces additional important benefits in comparison to these methods.

\smallskip \noindent \textbf{Automatic balancing between RL and IL policies}. First, IBRL does not require picking hyper-parameters nor annealing schedules for the BC regularization weight.
Unlike prior methods, IBRL does not need to worry about the IL policy being washed out in the early stage of training nor does it need to worry about the BC causing RL to be suboptimal in the later stage of training.
In IBRL, the balance between IL and RL changes automatically as the policy and critic improves.

\smallskip \noindent \textbf{Leveraging IL in both exploration and training.} The explicit consideration of IL actions during both exploration and training through the $\argmax$ operation (or $\softmax$ in the soft variant) can lead to better exploration and training targets when the RL policy is underperforming. We show later that both actor proposal and bootstrap proposal are crucial for maximum sample efficiency in ablations.

\smallskip \noindent \textbf{Modular and flexible architecture choices for IL and RL.} The modular design of IBRL easily enables selecting the ``best of both worlds'' from an IL and RL perspective. For example, we can use different network architectures that are most suited for the RL and IL tasks respectively. 
In \cref{sec:exp-robomimic}, we show that the widely used deep ResNet-18 encoder that achieves strong performance in IL performs poorly as the visual backbone for RL, while a shallow ViT encoder that performs worse in IL works quite well in RL.
IBRL's modular integration of RL and IL also allows different action representations for IL and RL, such as unimodal Gaussian for RL but mixture of Gaussians for IL.
This opens an avenue towards integrating some more powerful IL methods~\citep{reuss2023goal-diffuse, chi2023diffusionpolicy, zhao2023learning} with RL, which we leave for future research.

\subsection{Architectural Improvements}
\label{sec:design-choice}

\smallskip \noindent \textbf{Regularization with Actor Dropout.} 
Many prior works have demonstrated the benefit of regularization in RL for continuous control~\cite{fujimoto2018addressing, chen2021randomized, yarats2021image}. 
Additionally, as we discussed earlier, popular RL techniques that leverage prior data, such as oversampling demonstrations in training or adding BC regularization loss to policy update, implicitly introduce additional regularization to RL that has shown to be useful. 
We observe that regularizing IBRL with dropout~\citep{srivastava2014dropout} in the policy network (actor) $\pi_\theta$, which we refer to as \emph{actor dropout}, can further improve its stability and sample efficiency, especially in more challenging tasks where initial signals are noisy as successful episodes are less frequent.
Although dropout has been previously applied in the \emph{critic} to reduce overfitting on the value estimate~\cite{hiraoka2022dropout}, to the best of our knowledge, the application of dropout in \emph{actor} has not been well-studied before.
We find that adding actor dropout in IBRL significantly improves sample efficiency, even when other regularization techniques such as image augmentation (DrQ)~\citep{yarats2022mastering} or Q-ensembling (RED-Q)~\citep{chen2021randomized} are also present. 
Moreover, actor dropout accelerates convergence without increasing the update-to-data (UTD) ratio and requires negligible extra compute.


\begin{figure}[t]
    \centering
    \includegraphics[width=\linewidth]{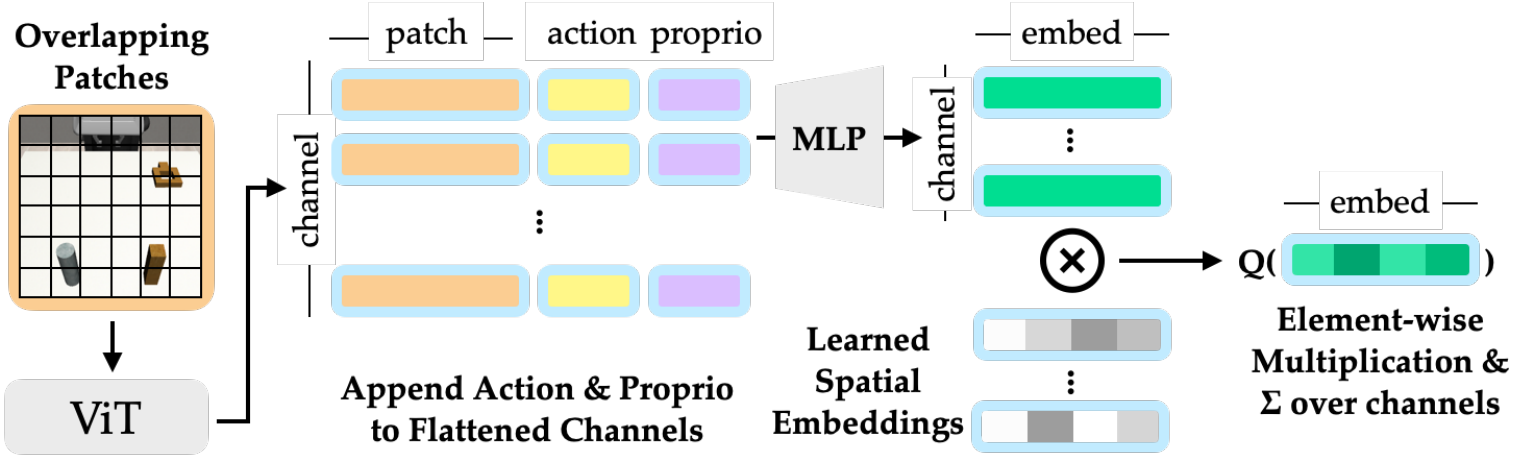}
    \caption{\small
    \textbf{ViT-based Q-network.} First, ViT processes overlapping image patches. Action and proprioception input are appended to each channel and an MLP is used to fuse this information. The projected embeddings are reduced to a 1-D vector by multiplying with learned spatial embeddings and summing over the channel dimension. Finally, an MLP takes the embedding and outputs a scalar $Q$. 
    }
    \vspace{-5mm}
    \label{fig:architecture}
\end{figure}

\smallskip \noindent \textbf{Improved Vision Encoder and Critic Designs.} 
Prior online RL in continuous control works have mostly inherited the architecture from DrQ~\cite{yarats2021image}, which consists of shallow ConvNet followed by linear layers.
Despite its strong performance in many settings, we find this architecture to be a major bottleneck in more challenging tasks.
Meanwhile, na\"ively applying common deep architectures without massive training data from parallel simulators leads to poor performance.
Therefore, we introduce a new Q-network design with a shallow ViT~\citep{dosovitskiy2020image} style image encoder for learning from pixels, illustrated in \cref{fig:architecture}. 
The general idea is to use Transformer layers so that relevant information from different parts of the image can be exchanged efficiently in a relatively shallow architecture that is expressive and yet easy to optimize. 
We first divide input images into \emph{overlapping} patches and apply two convolution layers to get patch embeddings before feeding them into one Transformer layer.
Then, we flatten the post-ViT patch embeddings in each channel and append the action and optionally proprioception data to each flattened channel before feeding them through an MLP to fuse this information.
To reduce dimensionality of the features without using large linear layers, we multiply the feature matrix with learned spatial embeddings~\cite{Kumar2022PreTrainingFR} and sum over the channel dimension to get a 1-D vector before feeding them into the final Q-MLP.
As TD3 utilizes two Q-heads for double Q-learning, we replicate the entire structure after the ViT for each Q-head. Similar to prior work, the actor is a fully connected network that takes the output of the ViT encoder as input.
We show that this architecture greatly improves the performance of IBRL in complex manipulation tasks in~\cref{sec:exp-robomimic} and show that it also improves baselines in the Appendix.

\section{Experiments in Simulation}


\label{sec:experiments}

\begin{figure*}[t]
\includegraphics[width=\linewidth]{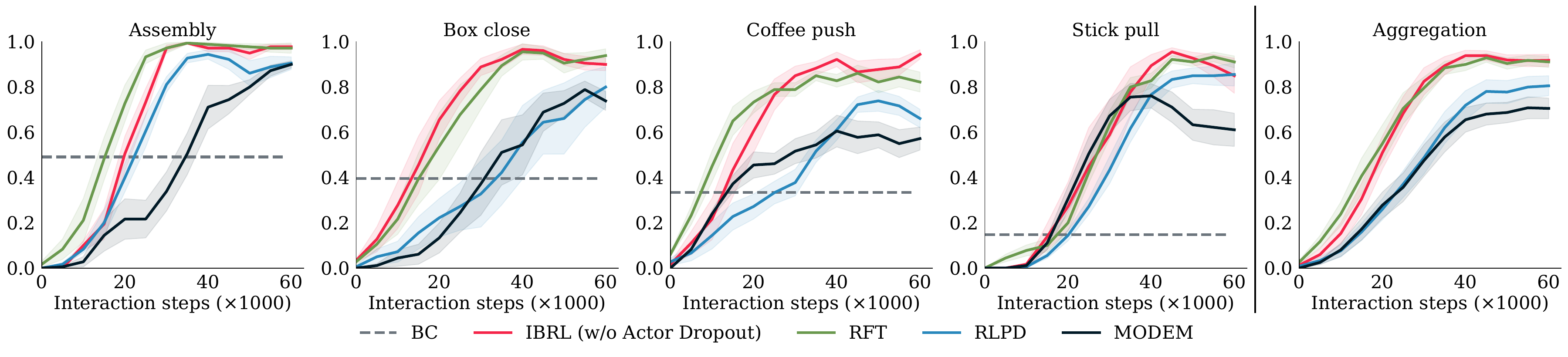}
\vspace{-4mm}
\caption{\small Performance on Meta-World.
IBRL (without Actor Dropout) outperforms both MoDem and RLPD on all 4 tasks. 
RFT achieves similar performance to IBRL.
The dashed lines indicate the average success rate of the BC policies used in IBRL.
}
\label{fig:metaworld_results}
\vspace{-4mm}
\end{figure*}

We first conduct experiments in simulation environments to comprehensively compare IBRL against state-of-the-art methods in terms of performance and sample efficiency. We also perform ablations to understand the importance of different design choices.

\subsection{Experimental Setup}

Our evaluation suite consists of 4 tasks from Meta-World~\citep{yu2019meta-world} and 2 tasks from Robomimic~\citep{mandlekar2021what}.
All environments use the sparse $0/1$ task completion reward at the end of each episode.
The 4 Meta-World tasks are a subset of the tasks evaluated in MoDem~\citep{hansen2023modem}. 
They span the medium, hard and very hard tiers of this benchmark as categorized in~\citep{seo2022masked}. 
Since Meta-World does not come with human demonstrations, we use the scripted expert policies from~\cite{yu2019meta-world} to generate 3 demonstrations per task.
Although we use harder-than-average tasks from Meta-World, these tasks are often simple, and additionally, scripted demonstrations are inevitably different -- much less noisy and cleaner -- than human demonstrations, making these tasks too simple to distinguish between some of the stronger methods. 
Robomimic is a well-established benchmark with significantly more complex tasks and demonstrations collected by human teleoperators. We use two test scenarios: a medium-difficulty task PickPlaceCan (Can) with 10 demonstrations and a hard task NutAssemblySquare (Square) with 50 demonstrations. 
As documented in~\cite{mandlekar2021what}, the Square task is particularly challenging for RL as RL methods without demonstrations have been unsuccessful even with hand-engineered dense rewards and substantial tuning.

\subsection{Implementation of IBRL and Baselines}
\label{sec:baselines}

\textbf{IBRL} uses TD3 for RL and BC for IL. The BC policies in all pixel-based experiments use a ResNet-18 vision encoder. 
We integrate common best practices for RL such as random-shift image augmentation in pixel-based RL and RED-Q in state-based RL to ensure best performance. 
Unless specified otherwise, IBRL always use actor dropout by default.
Please see the Appendix for more implementation details and a complete list of hyper-parameters.
We compare IBRL with three powerful baselines, RLPD, RFT and MoDem, that have been shown to outperform various other methods. 

\textbf{RLPD} loads the demonstrations in the replay buffer and oversamples them during online RL such that 50\% of the transitions in each batch come from demonstrations. 

\textbf{RFT} (regularized fine-tuning) is a technique where the RL policy $\pi$ is first pre-trained with demonstrations and then fine-tuned with online RL. During RL, it adds a BC loss $\alpha \lambda(\pi) L_\text{BC}$ where $\alpha$ is the weight of the BC loss and $\lambda$ is an annealing schedule. We use the soft Q-filtering technique from Regularized Optimal Transport (ROT)~\cite{haldar2022watch-rot} to dynamically anneal $\lambda$. We use the best $\alpha=0.1$ found through hyper-parameter sweeping.

RLPD and RFT share the same TD3 backbone as IBRL. In our experiments, unless otherwise specified, IBRL, RLPD, and RFT share the same non-algorithmic building blocks including network architecture, normalization, random-shift image augmentation, RED-Q, etc. We make these implementation decisions to ensure strong baselines and controlled comparisons against IBRL.

\textbf{MoDem} is a model-based approach that pre-trains a policy with BC and uses it to generate rollouts which are then used to pre-train a world model and critic. We use the original open-source implementation of MoDem. For our Meta-World experiments, we generate the prior demonstrations differently from the original paper \cite{hansen2023modem}, but we have confirmed that our rerun of MoDem with these demonstrations performs better on average than the results reported in the original paper.
 

\subsection{Overall Results on Meta-World and Robomimic}

\smallskip \noindent \textbf{IBRL matches or exceeds baselines in Meta-World.} 
In Meta-World, we focus on the core algorithmic contributions of IBRL. 
Therefore, we disable actor dropout for IBRL. We also \emph{do not} use our ViT-based architecture for IBRL, RFT, and RLDP but instead use the widely adopted ConvNet architecture from DrQ to ensure a fair comparison with MoDem as it is complicated to tune network architectures for MoDem.

\cref{fig:metaworld_results} shows the results of IBRL against three baselines in each Meta-World task separately as well as in aggregation (rightmost).
IBRL and RFT universally outperform RLPD and MoDem across all tasks in terms of both sample efficiency and final performance, solving all tasks within 40K samples. 
RFT has a small advantage over IBRL in the early stage of training thanks to its pretrained encoder and policy network. However, IBRL catches up quickly and achieves high performance within the same amount of samples, significantly outperforming RLPD which is also randomly initialized.
Because the tasks are relatively simple, the IBRL's advantage of integrating a more powerful IL model is less beneficial here, which may partially explain the similar performance between IBRL and RFT.
However, it is worth noting that RFT requires additional tuning to find a proper range for the base regularization ratio $\alpha$. In contrast, IBRL has no additional hyper-parameters during the RL stage, making it more desirable for real world applications where large scale hyper-parameter search is infeasible.
Lastly, the more complex MoDem method performs much worse than IBRL and the two simpler baselines. Given that MoDem's computational cost is significantly higher than the other two baselines (10 hours for MoDem vs. 1 hour for the three model-free methods), we exclude MoDem in the more difficult and computationally intensive Robomimic experiments.

\label{sec:exp-robomimic}

\begin{figure*}[t]
\includegraphics[width=\linewidth]{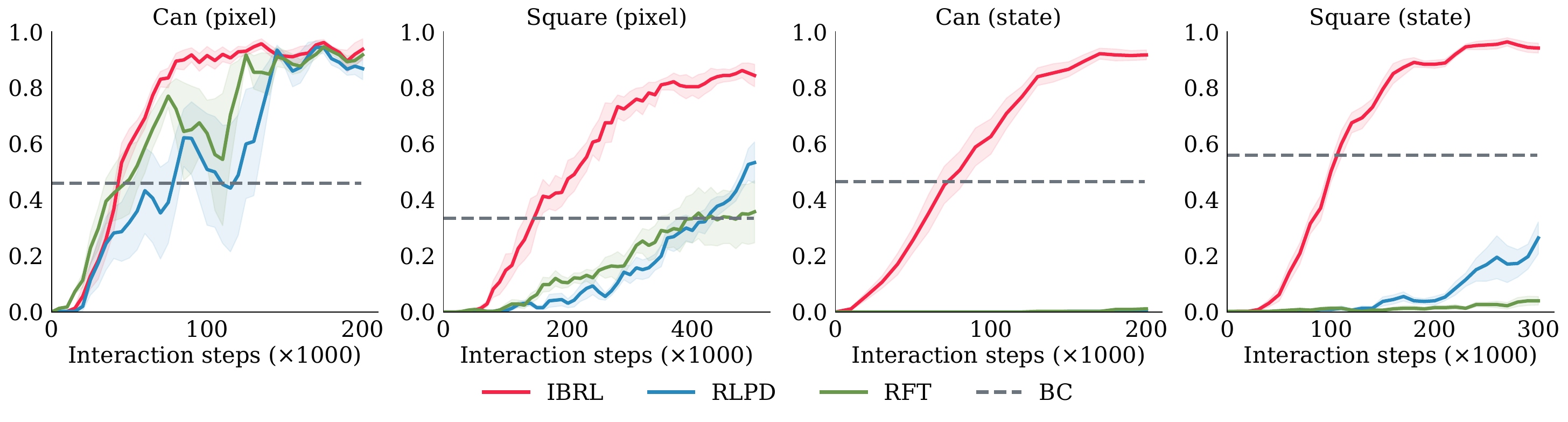}
\vspace{-4mm}
\caption{\small Performance on Robomimic.
IBRL significantly outperforms RFT and RLPD on all 4 scenarios. The gap between IBRL and baselines is especially large on the more difficult Square task.  
The horizontal dashed lines are the score of BC policies in IBRL.
}
\label{fig:robomimic_results}
\vspace{-4mm}
\end{figure*}

\smallskip \noindent \textbf{IBRL significantly exceeds baselines in Robomimic.} 
In Robomimic, we run all methods with our new ViT-based architecture as existing architectures become a major bottleneck in Square, the most complicated task in our simulation experiments. 
We also run state-based experiments to demonstrate the effectiveness of IBRL in isolation from network designs. 
We run IBRL with actor dropout to highlight our empirical improvement upon existing strong baselines. The ablations over different components of IBRL are in the next section.

~\cref{fig:robomimic_results} shows the performance of IBRL alongside the two strong baselines, RLPD and RFT. 
IBRL outperforms the baselines across all four settings. 
The performance of the BC policy (gray dashed lines) illustrates the relative difficulty of the tasks. 
For example, Square (pixel) is much harder than Can (pixel); BC performs much worse in Square despite having $5\times$ as much demonstration data as Can. 
In the relatively simpler Can (pixel) task, all three methods are able to eventually solve the task, but IBRL solves it with fewer interaction steps and more stable training. In the Square (pixel) task, IBRL is the only method that is able to solve it within 0.5M samples, while the baselines attain less than 60\% success. In state-based setting, the improvement is even more striking as the existing methods fail to learn completely. Overfitting may be a major issue that leads to the failures of baselines in state-based experiments as we later see that their performance improves significantly after adding actor dropout, despite still being worse than IBRL.

\subsection{Ablations on Robomimic}

We perform ablations on the more challenging Robomimic tasks to understand the contribution of each components of IBRL.
We first show that adding actor dropout to the baselines is not sufficient to match IBRL's performance. 
Then we ablate over the algorithmic components of IBRL and show that all of them contribute to its success. 
Finally, we show that our ViT-based architecture significantly improves sample efficiency and final performance for all RL methods.

\begin{figure}[t]
\includegraphics[width=\linewidth]{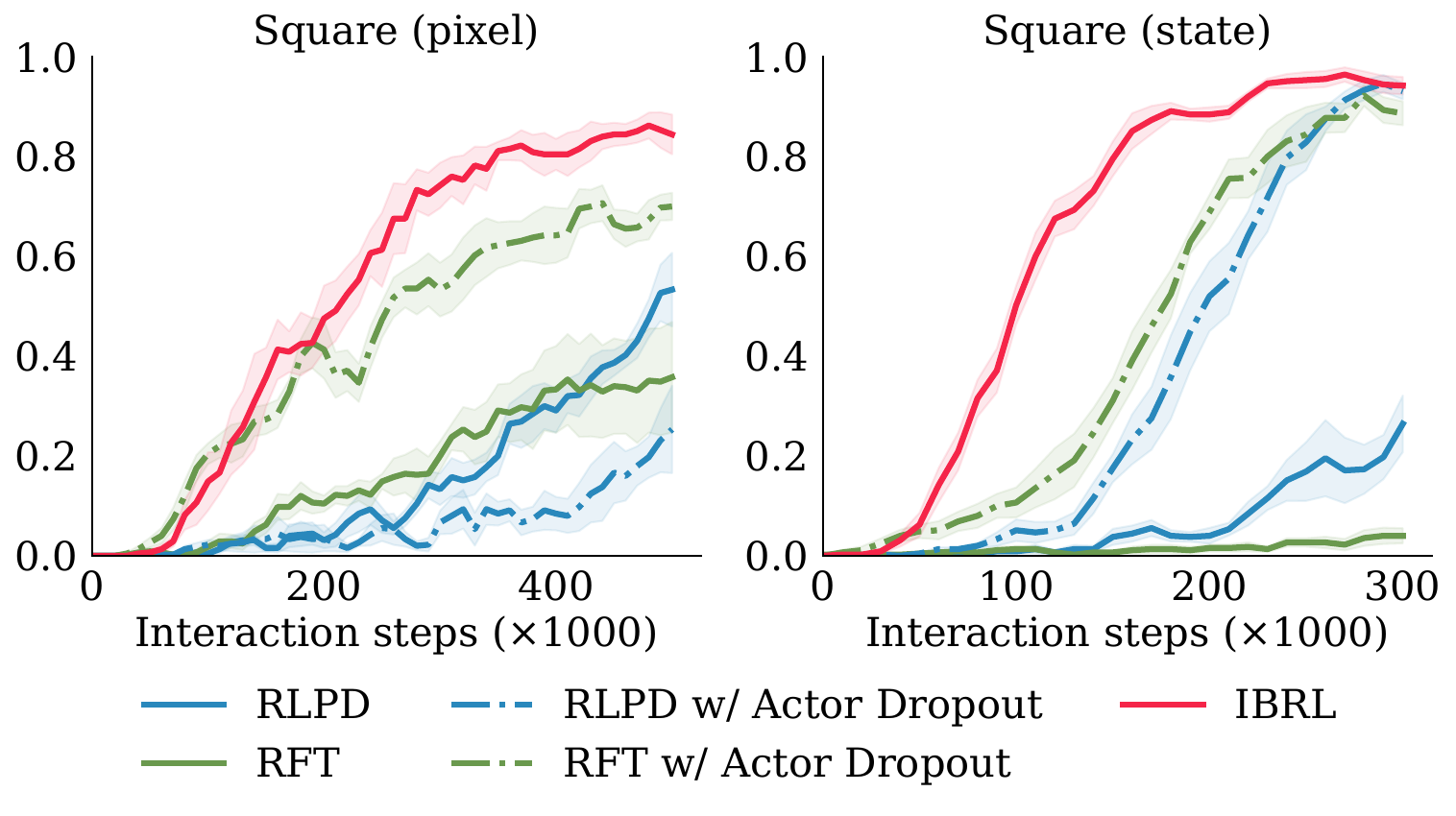}
\vspace{-4mm}
\caption{\small IBRL vs. baselines and their variants with actor dropout. Actor dropout significantly improves RFT in pixel-based RL and significantly improves both baselines in state-based RL. 
}
\label{fig:actor_dropout}
\vspace{-2mm}
\end{figure}

\smallskip \noindent \textbf{Actor dropout on baselines.} To ensure that the advantage of IBRL over the baselines are not solely from actor dropout, we augment both RLPD and RFT with actor dropout and show their performance in~\cref{fig:actor_dropout}. 
First of all, IBRL still outperforms the strongest variant among the four baselines, ``RFT with Actor Dropout", showing that actor dropout is not the only reason behind IBRL's new SoTA performance. 
However, it is worth noting that actor dropout significantly improves RFT in both pixel- and state-based settings and RLPD in state-based setting. In the state-based setting, actor dropout essentially helps the two baselines solve the task, although at a lower sample efficiency than IBRL.
Adding actor dropout to RFT essentially leads to a new approach that greatly surpasses existing methods excluding IBRL.
This suggests that this technique should be considered for other methods beyond IBRL, especially considering that it adds negligible extra computational cost.

\begin{figure}[t]
\includegraphics[width=\linewidth]{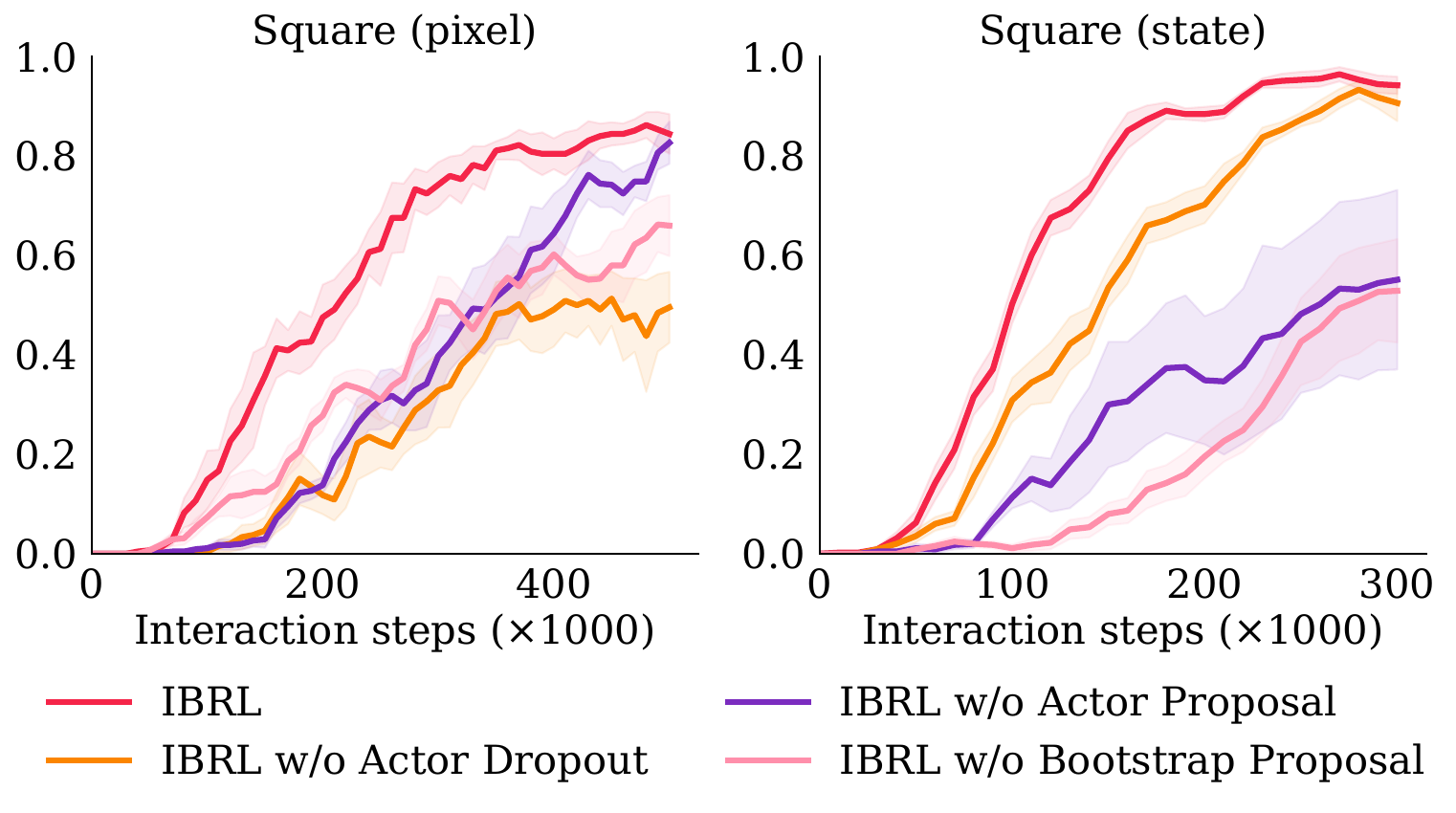}
\vspace{-4mm}
\caption{\small 
Ablations on the algorithmic components of IBRL.
}
\label{fig:ibrl_ablation}
\vspace{-4mm}
\end{figure}

\smallskip \noindent\textbf{Algorithmic components of IBRL.} 
To understand the importance of key algorithmic components in IBRL, we perform ablations over actor proposal, bootstrap proposal, and actor dropout in \cref{fig:ibrl_ablation}. 
Overall, all three components are crucial for IBRL's strong performance. 
First, we can see that actor dropout is a powerful technique that improves sample efficiency and helps IBRL to escape sub-optimal solutions. Nonetheless, we emphasize that the core ideas of IBRL play a crucial role even when actor dropout is enabled: removing either the bootstrap proposal or actor proposal causes significant performance deterioration even when actor dropout is enabled.
IBRL w/o Bootstrap Proposal shares a similar high-level structure to PEX ~\cite{zhang2023policy}, where a reference policy is used for proposing actions during exploration only. However, PEX trains the reference policy with offline RL and does not use actor dropout. 
IBRL is significantly less sample efficient without bootstrap proposal, indicating that using the IL policy in the target value computation leads to better training targets and faster convergence. 
We also verify the importance of the actor proposal; IBRL's performance decreases when removing actor proposal because it becomes less efficient at finding good actions in early stage of training. 
It is interesting to see that IBRL w/o Bootstrap Proposal performs worse than IBRL w/o Actor Proposal, which further emphasizes the importance of using the IL policy during training.

\begin{figure}[t]
\includegraphics[width=\linewidth]{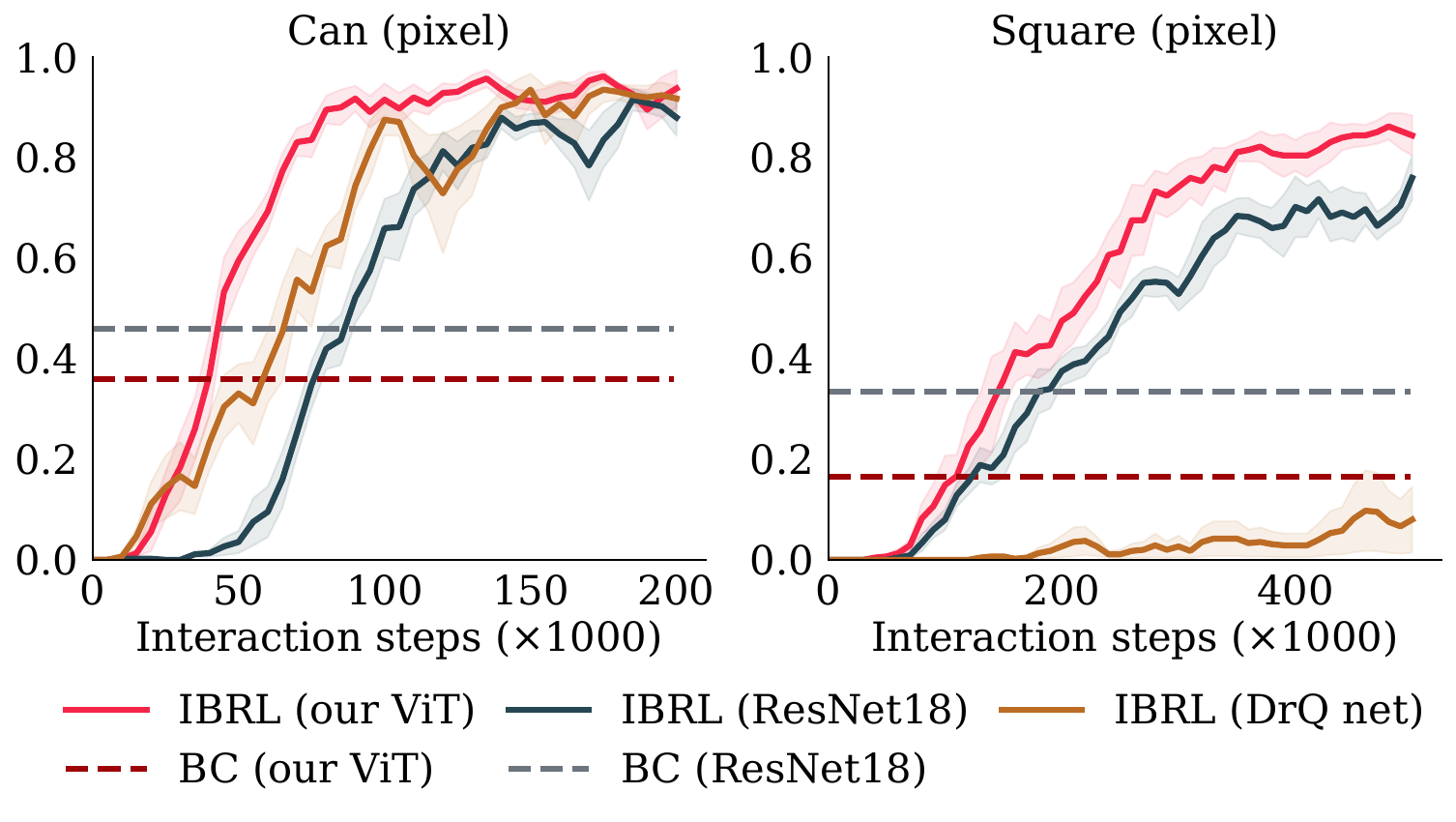}
\vspace{-4mm}
\caption{\small 
Comparison between our ViT-based Q-network design and the DrQ net commonly used in prior RL work and the ResNet-18 that achieves strong performance in imitation learning. 
All IBRL runs use the \emph{same} ResNet-18-based BC policy but different architectures for the RL networks. 
Dashed horizontal lines show the performance in BC. Our ViT performs significantly better in RL while the deeper ResNet-18 performs better in BC. IBRL takes advantage of the best architectures in both RL and IL.
}
\label{fig:ibrl_net}
\vspace{-4mm}
\end{figure}

\smallskip \noindent\textbf{Ablation of Network Architecture.} We demonstrate the effectiveness of our ViT-based architecture in~\cref{fig:ibrl_net}. In both tasks, our ViT architecture achieves better performance than the widely adopted DrQ network.
The near zero performance of DrQ network in Square also reflects the difficulty of the task compared to the ones used in prior RL works. 
We also test the deep ResNet-18 encoder, the same one used in our BC policy, in RL.
Note that this ResNet-18 replaces BatchNorm with GroupNorm~\cite{Wu_2018_groupnorm} as BatchNorm is known to cause RL to diverge when used with moving average target networks \cite{Kumar2022PreTrainingFR}.
Compared with the deeper and more computationally expensive ResNet-18, our proposed ViT architecture achieves better sample efficiency and final performance while also taking $50\%$ less wall-clock time to train. 
Although the ViT performs better in online RL tasks, we also see from~\cref{fig:ibrl_net} (dashed lines) that the higher capacity ResNet-18 still dominates in BC. 
Thus, we empirically confirm that BC and RL may prefer different architectures, which is reasonable given that the training goals are different (fitting behaviors in the training data vs. extrapolating to better behaviors while avoiding overfitting to unsuccessful early exploration data). 
Prior works such as RFT are forced to use the same architecture to fit both RL and demonstration data, which may limit their performance. 
In contrast, IBRL allows us to choose different architectures that are most suitable for RL and IL respectively, which echoes with one of the benefits of IBRL discussed in \cref{sec:benefits}.

\section{Real World Experiments}

To fulfill IBRL's promise of performing sample-efficient policy improvement in real-world applications, we evaluate it on three real-world manipulation tasks of increasing difficulty and compare it against RFT and RLPD.

\begin{figure}[]
\includegraphics[width=\linewidth]{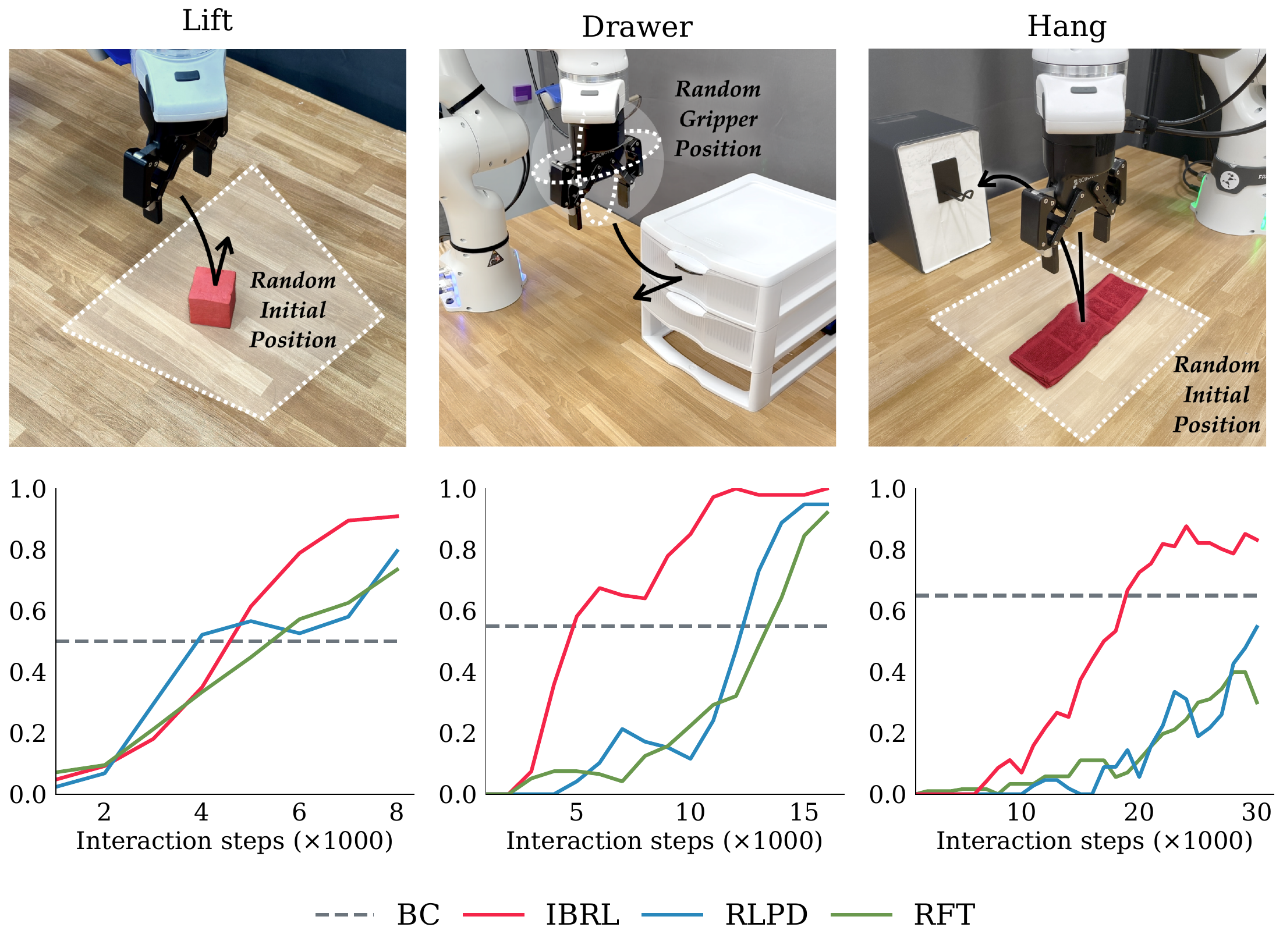}
\vspace{-4mm}
\caption{\small 
\textbf{Top:} Illustrations of each task and the variation in the initialization of each task.
\textbf{Bottom:} \emph{Training} curves for each task. $y$-axis is the percentage of successful episodes during each 1000-step interval. IBRL consistently outperforms RLPD and RFT in all 3 tasks, with a larger gap on the more complex tasks that take more interaction steps to learn. 
}
\label{fig:real_envs_and_learning_curves}
\vspace{-4mm}
\end{figure}

\subsection{Experimental Setup}

We design three tasks named Lift, Drawer and Hang. 
The first two tasks use a Franka Emika Panda robot and the third task uses a Franka Research 3 robot. 
Both robots are equipped with a Robotiq 2F-85 gripper. 
Actions are 7-dimensional consisting of 6 dimensions for end-effector position and orientation deltas under a Cartesian impedance controller and 1 dimension for absolute position of the gripper. Policies run at 10 Hz. 
For each task, we collect a small number of prior demonstrations via teleoperation with an Oculus VR controller, and then run different RL methods for a fixed number of interaction steps. 
All methods use the exact same hyper-parameters and network architectures as in Robomimic tasks.
We illustrate the three tasks in \cref{fig:real_envs_and_learning_curves} and briefly describe them here.

\smallskip \noindent \textbf{Lift:} The objective is to pick up a foam block. The initial location of the block is randomized over roughly 22cm by 22cm-28cm trapezoid, which covers the entire area visible from the wrist-camera when the robot is at the home position. We collect 10 demonstrations for this task due to its simplicity. It uses wrist-camera images as observations.

\smallskip \noindent \textbf{Drawer:} The objective is to open the top drawer in a set of plastic drawers in a fixed position. The initial pose of the robot is randomized by adding noise up to 10\% of the joint limit to each joint. We collect 30 prior demonstrations and use wrist-camera images as observations.

\smallskip \noindent \textbf{Hang:} The objective is to hang a deformable soft cloth on a metal hook. The initial location of the cloth is randomized over a roughly 28cm by 30cm rectangular region, and the hook is in a fixed position. The cloth is initialized so that its long side is roughly perpendicular to the hook. We use 30 prior demonstrations. This task uses third-person camera images as observations because the wrist-camera loses sight of the hook after picking up the cloth.

As the primary goal of our real-world evaluations is to compare sample-efficiency and performance of various algorithms, we design rule-based success detectors and perform manual reset between episodes to ensure accurate reward and initial conditions. The details of the success detection and reset mechanism are in the Appendix. 
Note that sparse $0/1$ reward from the success detector is the only source of reward.

\begin{table}[]
\centering
\begin{tabular}{l c c |c  c | c }
\toprule
&  \multirow{2}{*}{Lift} &  Lift         & \multirow{2}{*}{Drawer} & Drawer         & \multirow{2}{*}{Hang}\\
&                        & ({Hard Eval}) &                         & ({Early Stop}) & \\
\midrule
\# Demos      & 10             & 10    &   30 & 30      & 30    \\
BC            & 50\%           & 0\%   &   55\% & 55\% & 65\%     \\
\midrule
\# Env Steps  & 8K             & 8K    & 16K       & 10K  & 30K  \\
Time (mins) & 32  & 32 & 64  & 48 & 120 \\
RLPD          & 95\%           & 80\%  & 85\%      & 0\%  & 15\% \\
RFT           & 90\%           & 75\%  & 50\%      & 15\% & 35\% \\
IBRL          & \textbf{100\%} & \textbf{95\%}  & \textbf{95\%} & \textbf{100\%} & \textbf{85\%}\\
\bottomrule
\end{tabular}
\caption{Evaluation performance of IBRL on the real-world tasks.}
\label{tab:real_eval}
\end{table}

\begin{figure}
\centering
\includegraphics[width=\linewidth]{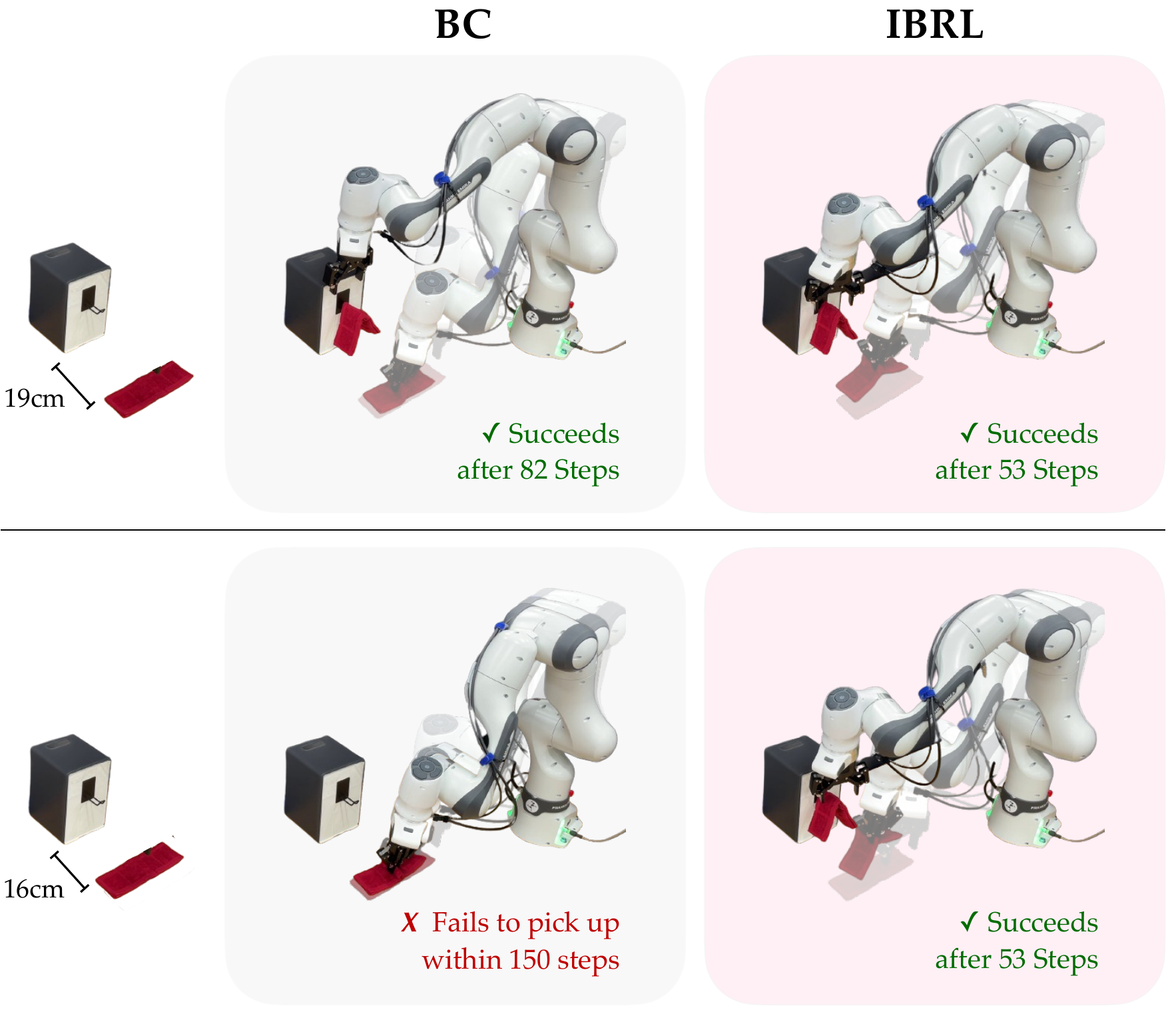}
\caption{Illustration of rollouts by BC and IBRL on the Hang task from two different initial cloth locations. Note that IBRL can achieve task success in fewer timesteps than the BC policy, and can solve the task for certain initial states where the BC policy fails.}
\label{fig:real-traj}
\vspace{-4mm}
\end{figure}

\subsection{Results}
~\cref{fig:real_envs_and_learning_curves} shows the training curves of IBRL and baselines in the three tasks. Different tasks allow different interaction budgets based on their difficulty. The training curve measures the success rate of episodes between each 1000-step interval while the policy is being updated and exploration noise for action is enabled. Overall, we see that IBRL learns consistently faster than RLPD and RFT across all three environments and is the only method that is able to outperform BC in the most challenging Hang task under a 30K interaction budget.

We take the last checkpoints of each method and perform 20 evaluations. All methods are evaluated using the same set of initial conditions for fairness. ~\cref{tab:real_eval} summarizes the results. 

In Lift, we first evaluate all methods using a uniform distribution of initial positions of the block and then evaluate them in a ``Hard Eval'' setting where the block is initialized at the boundary such that only part of the block is visible from the wrist-camera at the beginning of each episode. 
IBRL achieves the highest score in both settings. 
In the hard setting, performance of all methods decreases, especially for BC whose performance drops to $0$ as it has not seen such cases in the demonstrations. However, IBRL still maintains a near perfect $95\%$ success rate as it learns faster during RL and thus has seen more different initial positions. This illustrates that IBRL is highly suitable for real-world policy improvement to combat potential distribution shifts or to tackle unseen cases during original data collection.

The Drawer task is more challenging than Lift as it requires grasping of the small drawer handle followed by a precise horizontal motion to open the drawer. 
We provide 30 demonstrations and run each method for 16K interaction steps. IBRL achieves the strongest performance at 95\% success. 
From the learning curve in~\cref{fig:real_envs_and_learning_curves}, we can clearly see that IBRL solves the task with far fewer samples.
To verify this, we evaluate an ``early stop'' checkpoint after 10K interaction steps and find that IBRL already attains a perfect score while the baselines can only succeed in less than 15\% of the time. In fact, RLPD and RFT still cannot fully solve this task even after 16K steps, making IBRL at least $40\%$ more sample efficient than the baselines in this task.

The Hang task is the hardest task as the robot must learn to pick the cloth up from the center and release it above the hook with enough precision so that the cloth rests on the hook and does not fall. 
We provide 30 demonstrations and run each method for 30K interaction steps. BC performs relatively well on this task because the demonstrations from the human expert are clean and always grasp and drop at the optimal location, which reduces the possible state space that the policy needs to handle.
However, the deformable nature of the cloth makes it especially hard for RL as small differences in the grasp or drop locations may lead to drastically different outcomes that are hard to predict.
Despite a significantly higher online interaction budget of 30K steps, RFT and RLPD are not able to even reach the performance level of BC. In contrast, IBRL exceeds the success rate of BC by $20\%$. 
\cref{fig:real-traj} illustrates rollouts of BC and IBRL on two different initial conditions. 
In the top row, IBRL is able to solve the task with fewer steps than the BC policy. The bottom row shows a different scenario where BC fails to pick up the towel within the episode limit of 150 steps while IBRL can still solve the task.

\section{Discussion}
\smallskip \noindent \textbf{Summary.} We present IBRL, a novel way to use human demonstrations for sample efficient RL by first training an IL policy and using it in RL to propose actions to improve online interaction and training time target Q-value estimation.
We show that IBRL outperforms prior SoTA methods across 6 simulation tasks spanning wide range of difficulty levels and the improvement is particularly more significant in harder tasks.
In real-world robotics tasks, IBRL also outperforms prior methods by a large margin in terms of sample-efficiency and final performance, making it an ideal solution for rapid real-world policy improvement to either improve upon an existing IL policy or help address performance deterioration caused by distribution shift.
While we instantiated IBRL with specific choices of IL and RL algorithms, the framework is general and can in principle accommodate any IL method and off-policy RL method.

\smallskip \noindent \textbf{Limitations and Future Work. }
In our real-world experiments, we focus on evaluating the performance of IBRL so we resort to manual reset to minimize noise from unsuccessful resets. A large scale deployment of IBRL in the real world should ideally enable autonomous reset, which we leave for future work.
Additionally, the modular design of IBRL opens new avenues for integrating various IL methods with RL. An exciting direction for future research is to extend IBRL to take advantage of recent IL advancements such as diffusion policies \citep{reuss2023goal-diffuse, chi2023diffusionpolicy} or learning with hybrid actions~\citep{belkhale2023hydra} for even better performance.

\section{Acknowledgments}
This project was sponsored by ONR grant N00014-21-1-2298, NSF grants \#2125511, \#1941722, \#2006388 and DARPA grant W911NF2210214. We would like to thank Yuchen Cui, Joey Hejna for their feedbacks and suggestions.

\bibliographystyle{plainnat}
\bibliography{references}

{
\onecolumn

\section{Pseudocode for IBRL}

\label{sec:ibrl-code}
\begin{algorithm}[h]
\caption{IBRL with TD3 backbone. Major modifications w.r.t. vanilla TD3 highlighted in {\color{RoyalBlue} blue}.}
\label{algo:ibrl}
\begin{algorithmic}[1]
\State \textbf{Hyperparameters}: number of critics $E$, number of critic updates $G$, update freq $U$, exploration std $\sigma$, noise clip $c$
\State {\color{RoyalBlue} Train imitation policy $\mu_{\psi}$ on demonstrations $\mathcal{D} = \{{\xi_1, \dots, \xi_n}\}$ with the selected IL algorithm.}
\State Initialize policy $\pi_\theta$, target policy $\pi_{\theta'}$, and critics $Q_{\phi_i}$, target critics $Q_{\phi_i'}$ for $i = 1, 2,\dots, E$
\State {\color{RoyalBlue} Initialize replay buffer $B$ with demonstrations $\{{\xi_1, \dots, \xi_n}\}$}
\For {$t = 1, \dots, \text{num\_rl\_steps}$}
  \State Observe $s_t$ from the environment
  \State {\color{RoyalBlue} Compute IL action $\ail_t \sim \mu_{\psi}(s_t)$} and RL action $\arl_t = \pi_\theta(s_t) + \epsilon$, $\epsilon \sim \mathcal{N}(0, \sigma^2)$
  \State Sample a set $\mathcal{K}$ of $2$ indices from $\{1, 2, \dots, E\}$
  \State {\color{RoyalBlue} Take action with higher Q-value  $a_t = \argmax_{a\in \{\arl, \ail\}} [\min_{i \in \mathcal{K}}Q_{\phi_i'}(s_t, a)]$}
  \State Store transition $(s_t, a_t, r_t, s_{t+1})$ in $B$
  \If {$t ~\% ~U \neq 0$}
    \State Continue
  \EndIf
  \For{$g = 1, \dots, G$}
    \State Sample a minibatch of $N$ transitions $(s^{(j)}_t, a^{(j)}_t, r^{(j)}_t, s^{(j)}_{t+1})$ from $B$
    \State Sample a set $\mathcal{K}$ of $2$ indices from $\{1, 2, \dots, E\}$
    \State For each element $j$ in the minibatch, compute target Q-value
    $$ 
    y^{(j)} = r_t^{(j)} + \gamma {\color{RoyalBlue} \max_{a' \in \{\ail, \arl\}}}
        \left[\min_{i \in \mathcal{K}}Q_{\phi_i'}(s_{t+1}, a')\right]
    $$
    $$
    {\color{RoyalBlue} \ail \sim \mu_\psi(s_{t+1})} ~\text{and}~\arl = \pi_{\theta'}(s_{t+1}) + \text{clip}(\epsilon, -c, c)
    $$
    \State Update $\phi_i$ by minimizing loss: 
        $L(\phi_{i}) = \frac{1}{N} \sum_j [y^{(j)} - Q_{\phi_i}(s^{(j)}_{t}, a^{(j)}_{t})]^2$ 
        for $i = 1, \dots, E$
    \State Update target critics $\phi_i' \leftarrow \rho\phi_i' + (1-\rho)\phi_i$ 
        for $i = 1, \dots, E$
    \EndFor
    \State Update $\theta$ with the last minibatch by \emph{maximizing}
    $\frac{1}{N} \sum_{j} \min_{i=1,\dots,E}Q_{\phi_{i}}(s^{(j)}_t, \pi_\theta(s^{(j)}_t))$
    \State Update target actor $\theta' \leftarrow \rho\theta' + (1-\rho)\theta$
    \EndFor
\end{algorithmic}
\end{algorithm}

\cref{algo:ibrl} contains the detailed pseudocode for IBRL.
Lines 2-4 do the necessary initialization for policy, critics and replay buffer. 
Then lines 6-10 correspond to interacting with the environment and line 9 specifically corresponds to the \emph{actor proposal} of IBRL. 
Note that the minimization over the multiple critics $\min_{i \in \mathcal{K}}Q_{\phi_i'}(s_t, a)$ is part of TD3 and RED-Q.
Lines 12-16 are critic updates and line 14 is the \emph{bootstrap proposal}. Finally, lines 18-19 are policy updates, which is identical to vanilla TD3. The final output of IBRL is the hybrid policy that acts following  $a_t = \argmax_{a\in \{\arl, \ail\}} [\min_{i \in \mathcal{K}}Q_{\phi_i'}(s_t, a)]$ .
The code shown here uses the $\argmax$ action selection scheme, we can obtain the $\softmax$ version by simply replacing the action selection method in line 9 and line 14 with
$a \sim \softmax_{a\in \{\ail, \arl\}}(\beta Q(a))$.

\section{Implementation Details and Hyperparameters}
\label{app:impl}

\begin{table}[]
\centering
\bgroup
\def\arraystretch{1.5}
\begin{tabularx}{0.74\textwidth}{p{125pt}c|c|c | c}
\toprule
Parameter & Meta-World &  Robomimic (Pixel) & Real-World &  Robomimic (State) \\
\midrule
Optimizer          & \multicolumn{4}{c}{Adam} \\
Learning Rate      & \multicolumn{4}{c}{$1\mathrm{e}{-4}$} \\
Batch Size         & \multicolumn{4}{c}{256} \\
Discount ($\gamma$) & \multicolumn{4}{c}{0.99} \\
Exploration Std. ($\sigma$) & \multicolumn{4}{c}{0.1} \\
Noise Clip ($c$) & \multicolumn{4}{c}{0.3} \\
EMA Update Factor ($\rho$) & \multicolumn{4}{c}{0.99} \\
Update Frequency ($U$)  &  \multicolumn{4}{c}{2} \\
Actor Dropout & \multicolumn{4}{c}{0.5} \\
Q-Ensemble Size ($E$)    & \multicolumn{3}{c|}{2}  &  5  \\
Num Critic Update ($G$)  & \multicolumn{3}{c|}{1}  &  5  \\
Inverse Temperature (Soft-IBRL, $\beta$) & \multicolumn{3}{c|}{N/A} & 10 \\
\midrule
Image Size         & $84 \times 84 $ & \multicolumn{2}{c|}{$96 \times 96$} & N/A\\
Use Proprio        & No              & \multicolumn{2}{c|}{Yes} & N/A \\
Proprio Stack      & N/A              & \multicolumn{2}{c|}{3} & N/A \\
State Stack & \multicolumn{3}{c|}{N/A}   & 3 \\
Action Repeat      & 2               & \multicolumn{3}{c}{1} \\
\bottomrule
\end{tabularx}
\caption{Hyperparameters for IBRL.}
\label{tab:hparam}
\egroup
\vspace{-5mm}
\end{table}

In this section we cover the implementation details of IBRL as well as the baselines.

The BC policies use a ResNet-18 encoder. The output of the ResNet encoder is flattened and then fed into the MLPs to get the final $7D$ actions.
For all the ResNet encoders used in this paper, we replace the BatchNorm layers in ResNet with GroupNorm~\citep{Wu_2018_groupnorm} and set the number of groups equal to the number of input channels. The modified ResNet achieves similar performance as the original one in BC but significantly better in RL since BatchNorm does not work well with exponential moving average target networks in RL.
We train the BC policies using Adam optimizer~\citep{kingma2014adam} with batch size of $256$ and learning rate of $1\mathrm{e}{-4}$. 
We use random-shift data augmentation to prevent overfitting. 
In Meta-World, we follow the camera position used in MoDem~\cite{hansen2023modem} for fair comparison.
Prior work~\cite{hsu2022visionbased} shows that wrist cameras improve generalization and sample efficiency. Therefore, we opt for wrist cameras whenever possible in Robomimic and real-world experiments. Specifically, we use the wrist camera in Can (Robomimic), Lift (real-world) and Drawer (real-world). In Square (Robomimic) and Hang (real-world), we use the 3rd-person camera because the wrist camera may not capture the goal location in this task.
In Robomimic, we additionally experiment with state-based IBRL where the BC policies use a straightforward 4-layer MLP with 1024 hidden units per layer.
The input to the policy is the stack of three states at $t$, $t\!-\!1$ and $t\!-\!2$. We find that MLPs with stacked state inputs achieve similar performance as the LSTMs from~\cite{mandlekar2021what}.
We use dropout $0.5$ in state-based BC to prevent overfitting.

The major hyperparameters for RL in IBRL are listed in \cref{tab:hparam}.
In pixel-based RL, the RL policies use the same camera view as the BC policies in each environment. Following DrQ-v2~\cite{yarats2022mastering}, the actor and two critics share the image encoder but only the gradients from the critics are used to update the image encoder.
We also use random-shift data augmentation in RL to  prevent overfitting and improve sample efficiency.
Different from~\cite{yarats2022mastering} which only uses target networks for critics, we also use a target actor as we find it slightly improves training stability.
In environments that use proprioception data, we use a stack of three proprioception data $(t, t\!-\!1, t\!-\!2)$ instead of only using the current proprioception data ($t$). The details of our ViT-based architecture are shown in ~\cref{fig:vit-code}, \cref{fig:critic-code} and \cref{fig:actor-code}.
In state-based RL, we use Q-ensembling (RED-Q) with $E=5$ and a higher UTD ratio $G=5$ as we find this combination achieves good sample efficiency. We have also tried to further increase UTD ratio to $G=10$ but find it takes significantly longer wall-clock time to train without improving sample efficiency. Critics and the actor in state-based RL are all 4-layer MLPs shown in~\cref{fig:state-code}.
Similar to state-based BC, we use a stack of three states as the input for critics and the actor.
We set actor dropout with $p = 0.5$ in all environments.
In Meta-World, we inherit the action repeat value from prior work for fair comparison. We do not use action repeat for Robomimic and real-world tasks.

The RLPD and RFT baselines share the same base RL implementation as IBRL. The core idea of RLPD~\cite{ball2023efficient} is to draw half of the batch from demonstrations and the other half from the RL replay buffer to upweigh the successful demonstration trajectories to address the exploration challenge. Note that the original RLPD paper use SAC as the base RL algorithm while our implementation use the same TD3 as IBRL for controlled experiments. Note that the original RLPD disables the entropy backup part of SAC in 3 out of 4 benchmarks evaluated, making that specific SAC variant highly similar to TD3 in practice. 

RFT first pretrains the encoder and policy head with BC and then runs RL with an additional BC loss term on the policy head for regularization. Different instantiations of this idea have appeared in prior works~\cite{nair2018overcoming}. Specifically, our implementation of RFT resembles the one from ROT~\cite{haldar2022watch-rot}. The policy loss is
$L_\theta(\pi_{\theta}) = -E_{s \sim \mathcal{D}}Q(s, \pi_\theta(s)) + \alpha \lambda(\pi_\theta) E_{(s, a)\sim \mathcal{T}} \| a - \pi_\theta(s) \|^2 $, where $\mathcal{D}$ is the RL replay buffer and $\mathcal{T}$ is the demonstration dataset. Moreover, $\alpha$ is the base regularization weight and we set $\lambda(\pi_\theta) = E_{s\sim \mathcal{T}} [\mathbbm{1}_{Q(s, \pi^0_\theta(s)) > Q(s, \pi_\theta(s))}]$ to dynamically adjust the weight of regularization. $\pi^0_\theta(s)$ is the pretrained policy.

\section{ViT-based Architecture Improves All Methods}

\begin{figure}[h]
\centering
\includegraphics[width=\linewidth]{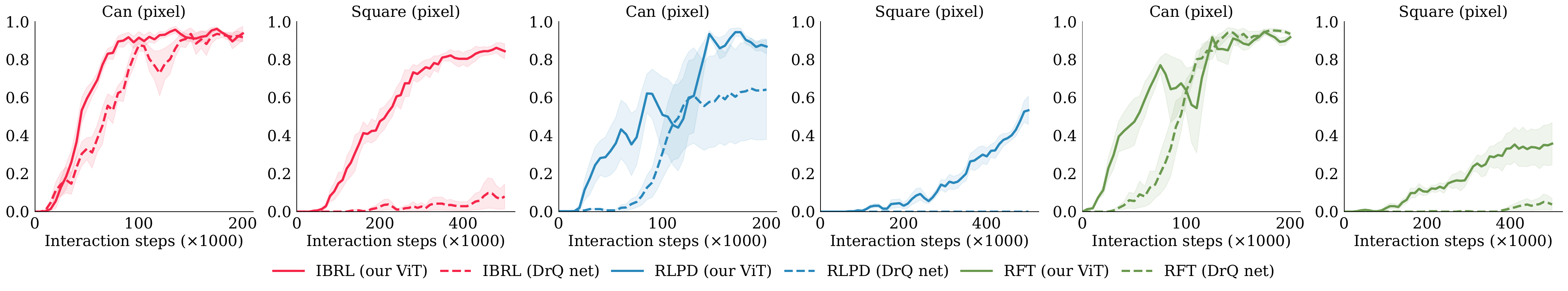}
\caption{Performance of our ViT v.s DrQ network on IBRL, RLPD and RFT. Our ViT-based architecture universally improve all three methods.}
\label{fig:vit-perf}
\end{figure}

In the main paper we show that our ViT architecture improves performance for IBRL over the commonly used network architecture from DrQ~\cite{yarats2022mastering}. 
To show that the improvement from our ViT architecture is general, we additionally evaluate RLPD and RFT on both our ViT and DrQ network. 
~\cref{fig:vit-perf} summarizes the results. We clearly see that our ViT architecture greatly improves the performance of all three methods. This emphasizes that our contribution to a better network architecture is general and can be considered independent of IBRL in future work.

\section{Additional Details of Real-World Experiments}

\subsection{Success Detection}

We design rule-based systems to detect success of each task and give the final $0/1$ reward for each episode. 
We run each episode for a maximum number of steps depending on the time it requires to finish the task. 
An episode ends early when a success is detected.

\smallskip \noindent \textbf{Lift:} The objective is to pick up a foam block. We detect whether the gripper is holding the block by checking if the gripper width is static and the desired gripper width is smaller than the actual gripper width. 
The success detector returns 1 if the end effector has move upward by at least $2$cm while holding the block.
The maximum episode length is $75$.

\smallskip \noindent \textbf{Drawer:} The objective is to open the top drawer in a set of plastic drawers in a fixed position. 
We attach a red patch to the side of the drawer and install a side camera that detects the red patch. 
The red patch is visible to the side camera when the drawer is open and invisible when the drawer is closed.
The maximum episode length is $150$.

\smallskip \noindent \textbf{Hang:} The objective is to hang a deformable soft cloth on a metal hook. 
The cloth is the only red object in the scene so we can track its location. The success returns 1 when the gripper is wide open, the red pixels are stable, and highest red pixel is above a threshold.
The maximum episode legnth is $150$.

\subsection{Reset}

For \textbf{Drawer}, the reset is straightforward as we manually close the drawer if it is not fully closed. The robot will sample a new random initial location for the end effector at the beginning of each episode. For \textbf{Lift} and \textbf{Hang}, we follow a common reset strategy for all methods. At the beginning of training, we put the object in the center of the initial area and do not move it until the RL policy obtains its first success. If the object is moved before the first success, we put it back to the center. After the RL policy succeeds for the first time, we gradually move the object from the center to the boundary in each reset. If we reach the boundary before the training ends, we start resetting the object from top left to bottom right and repeat until training terminates.

\subsection{Safety Boundaries}

To prevent the robot from damaging itself and the scene, we set a safety boundary on the end effector position and rotation for each task. The boundary is by first getting the range of the end effector position and rotation from the human demonstrations and increasing the range by a fixed amount to get a modestly larger range for RL. For Hang, we additionally block the region right beneath the hook as the robot arm collides with the metal hook when the end effector is in that region. The episode terminates early with $0$ reward when the safety boundary is violated.

\section{Ablation of Actor Dropout on Meta-World}
\label{app:actor-dropout-mw}

\begin{figure}[t]
\centering
\includegraphics[width=\linewidth]{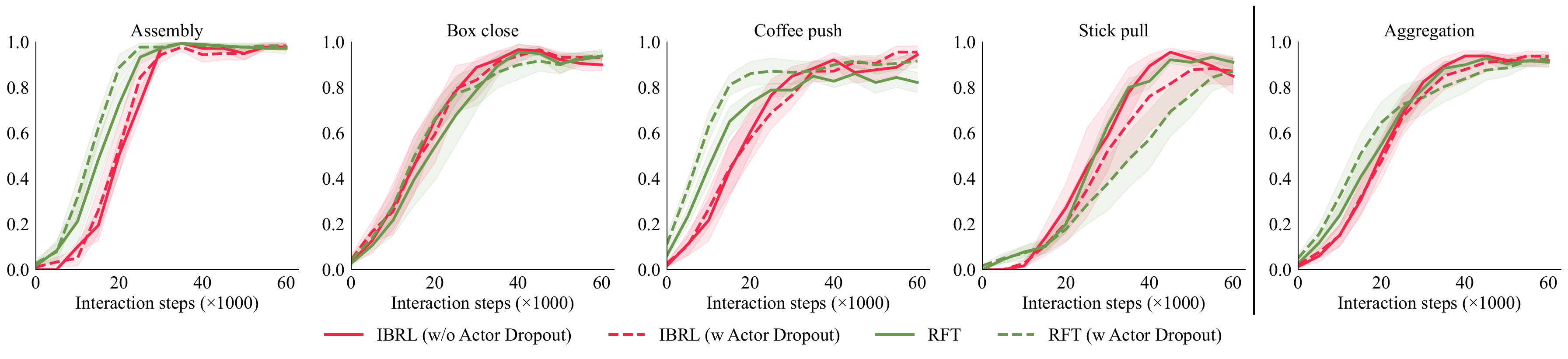}
\caption{Performance of RFT with actor dropout compared with IBRL counterparts. Actor dropout does not meaningfully change the performance of these strong methods on the relatively simple Meta-World benchmark.}
\label{fig:metaworld-dropout1}
\vspace{-2mm}
\end{figure}

\begin{figure}[t]
\centering
\includegraphics[width=\linewidth]{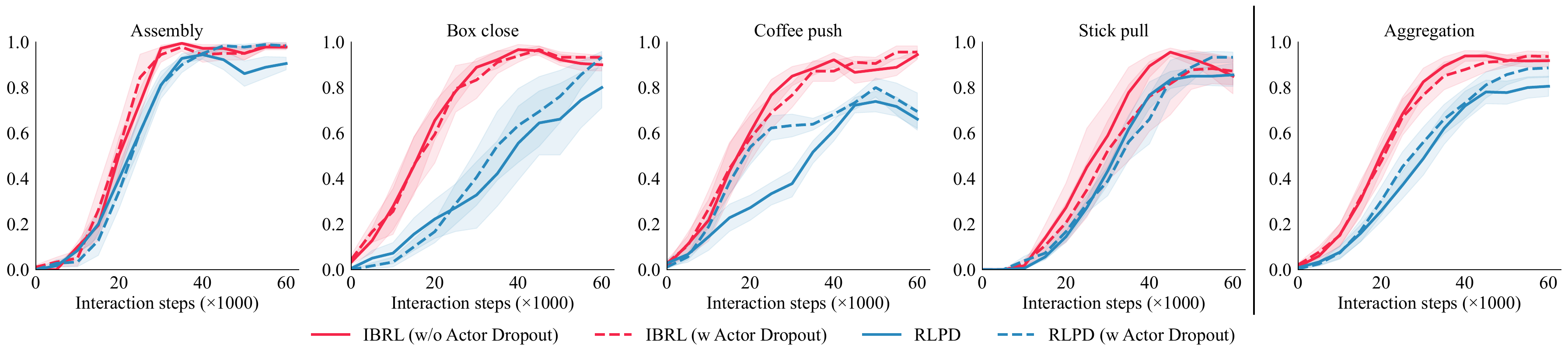}
\caption{Performance of RLPD with actor dropout compared with IBRL counterparts. Actor dropout slightly improves RLPD.}
\label{fig:metaworld-dropout2}
\vspace{-4mm}
\end{figure}

\begin{figure}[t]
\centering
\includegraphics[width=\linewidth]{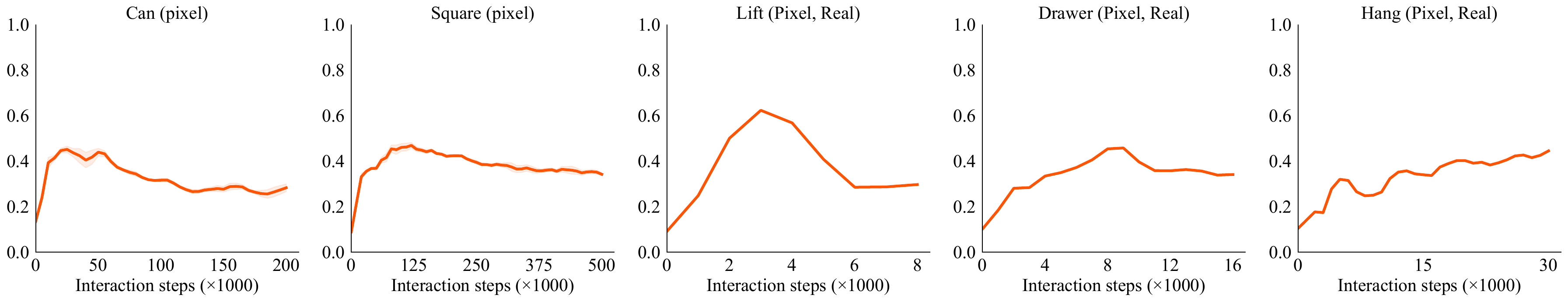}
\caption{Percentage of actions from BC policy selected during IBRL training.}
\label{fig:action-selection}
\vspace{-2mm}
\end{figure}

\begin{figure}[t]
  \centering
  \begin{minipage}[b]{0.48\linewidth}
    \centering
    \includegraphics[width=\linewidth]{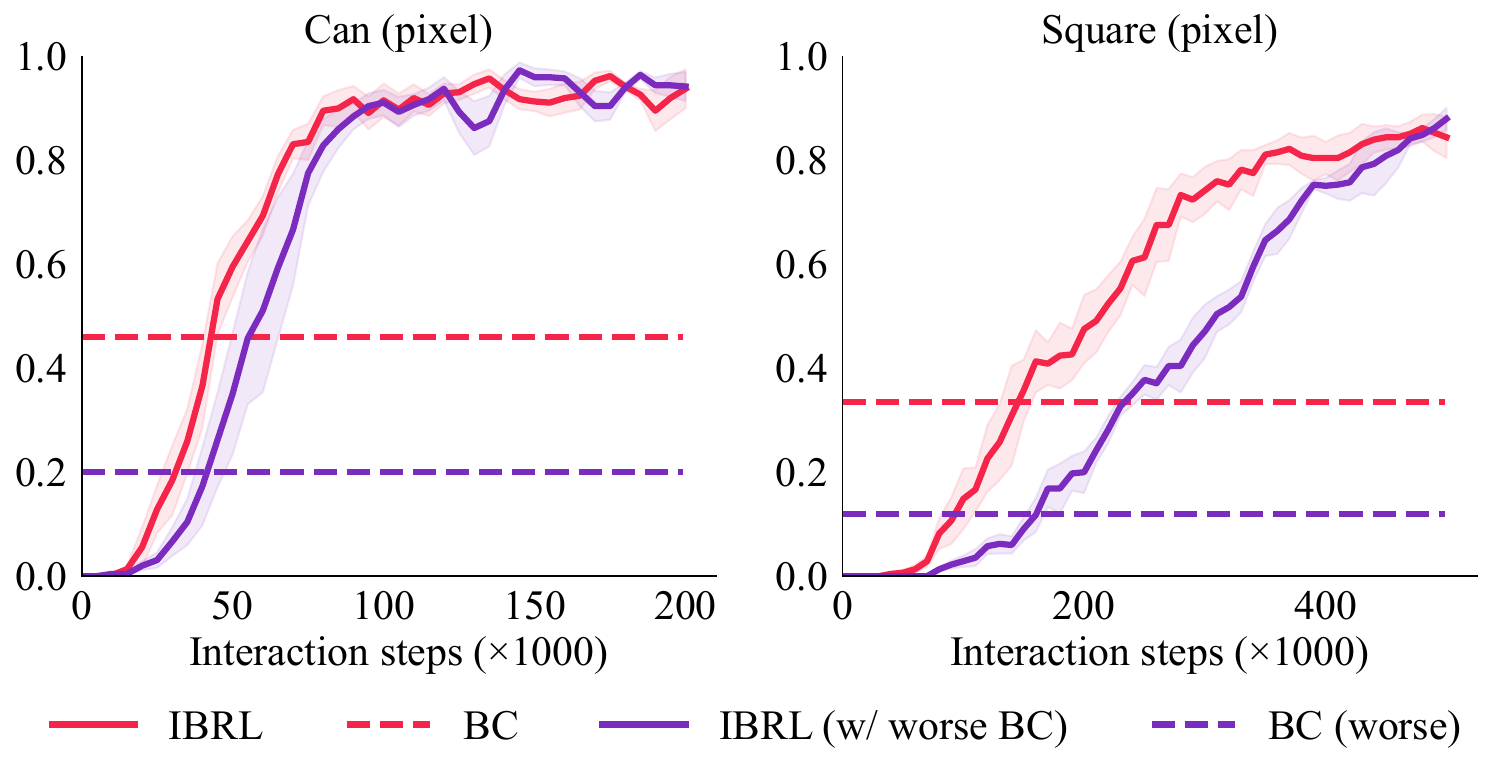}
    \caption{IBRL with a significantly worse BC policy trained from suboptimal demonstrations. This illustrates that IBRL can escape from substantially worse BC policies and achieve similar final performance at the cost of lower initial performance.}
    \label{fig:worse-bc}
  \end{minipage}
  \quad
  \begin{minipage}[b]{0.48\linewidth}
    \centering
    \includegraphics[width=\linewidth]{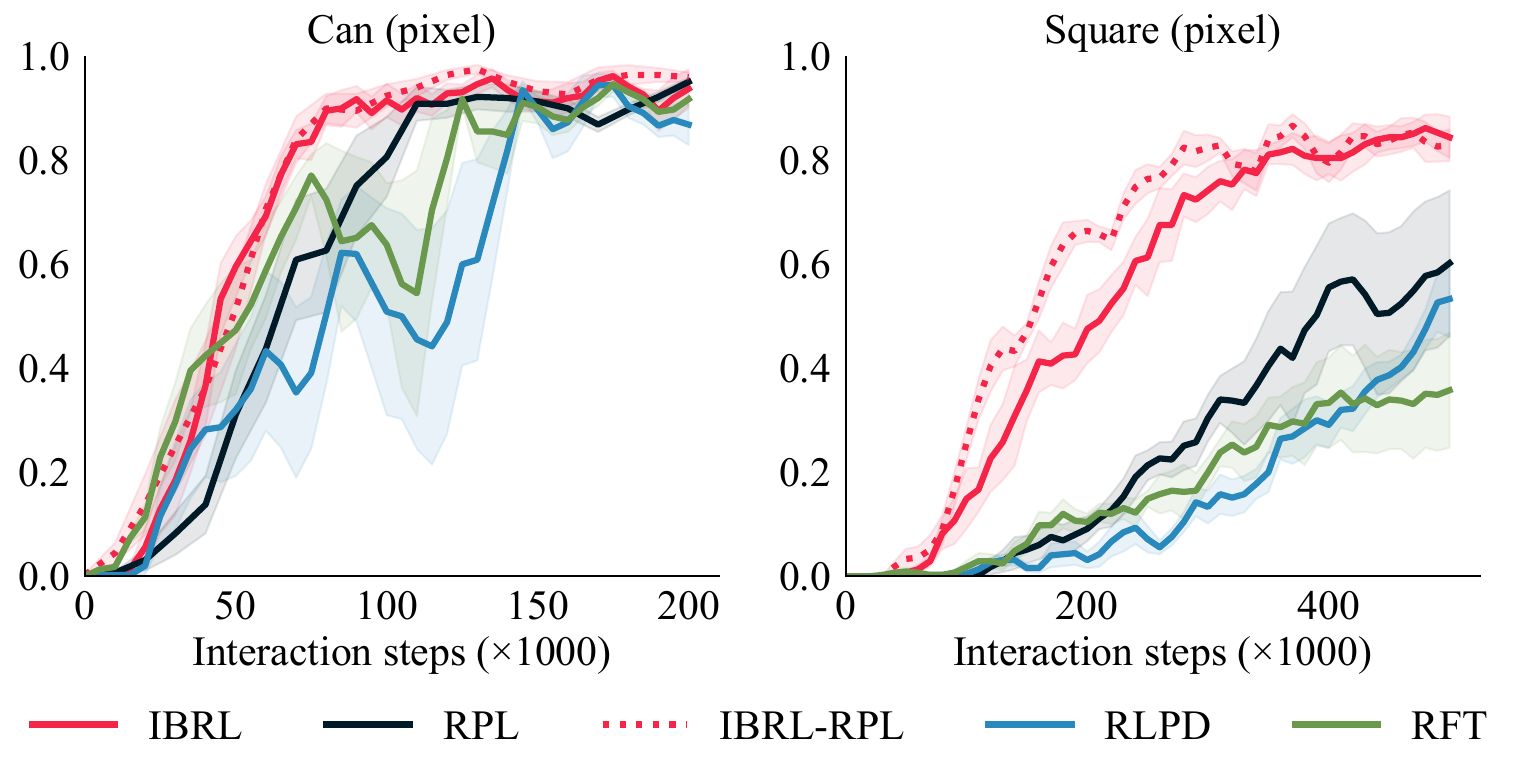}
\caption{Performance of Residual Policy Learning (RPL). RPL performs well among the baselines but still underperforms IBRL. RPL can be combined with IBRL to further improve performance on the harder Square task.}
\label{fig:rpl}
\end{minipage}
\vspace{-4mm}
\end{figure}

In the main paper, we do not include actor dropout in all methods on Meta-World as it does not meaningfully affect the conclusion in this simple benchmark. \cref{fig:metaworld-dropout1} and \cref{fig:metaworld-dropout2} shows the performance of IBRL, RFT and RLPD with and without actor dropout. Actor dropout slightly improves RLPD but makes little difference for IBRL and RFT which are already highly competitive in this benchmark.

Given these results, we want to emphasize that Meta-World is a relatively simple benchmark for single task RL as it is originally proposed for meta-learning and thus the designers ensure that each individual task can be solved easily~\cite{yu2019meta-world}. Specifically, Meta-World has smaller actions space (4 dimensional instead of 7 dimensional in Robomimic and real world), shorter episode length (less than 100 steps) and limited randomness in the initial condition. In the sparse reward setting considered in this paper, we observe that Meta-World tasks are easy for modern RL methods that utilize demonstrations even under highly limited data (i.e., 3 episodes of demonstrations). Therefore, these tasks do not provide enough signal to differentiate strong methods like IBRL and RFT, nor to justify the benefit of regularization techniques like actor dropout.

\section{Discussion on the IL policy in IBRL}
\label{app:il-in-ibrl}


To understand the role of the IL policy during IBRL training and at convergence, in \cref{fig:action-selection} we plot the frequency that IBRL selects the IL action when collecting online data for training. IBRL selects fewer actions from the IL policy at the beginning, because the critics are randomly initialized and it is easy for the RL actor to find actions with ``fake" high Q-values. As the critics get updated, the incorrectly high Q-values for those actions are pushed down and IBRL starts to pick more actions from IL policy as the critic learns that those IL actions are high quality by learning from the demonstration data in the replay buffer. Then, the ratio of IL actions steadily decreases in most cases as the RL policy improves. One exception is in the hardest real world Hang task, where the ratio of IL actions keeps increasing.
This is reasonable given that the IL policy is fairly strong for this task and the RL policy likely has not fully converged yet, as reflected by the high performance of IL and imperfect performance of IBRL in this task. In all cases, however, the ratio never decreases to zero, indicating that IBRL still relies on the IL policies, which are parameterized by much deeper networks, even at convergence.

Next, we investigate how a suboptimal IL policy may affect the performance of IBRL. We train suboptimal BC policies using the Multi-Human (MH) version of the Robomimic dataset instead of the Proficient Human (PH) version used in normal IBRL. The average length of the 50 demonstrations in the PH dataset is 149 compared to 271 in the MH dataset, indicating that the MH dataset comprises very inefficient motions. The dashed horizontal lines in ~\cref{fig:worse-bc} illustrate the performance gap of the BC trained from different dataset. The \emph{BC (worse)} policies achieve less than half of the success rates achieved by their counterparts trained on the PH data. We then run IBRL with \emph{BC (worse)} as the IL policy and keep everything else the same---i.e. we still add the PH data to the RL replay buffer for controlled experiments. As expected, the performance of IBRL decreases as the worse IL policies are unable to provide equally good alternative actions. However, IBRL is able to eventually escape from the worse BC to reach equally good final policies.

\section{Additional Baseline: Residual Policy Learning}
\label{app:rpl}

In this section, we compare IBRL with an additional baseline, residual policy learning (RPL)~\cite{rpl}. The core concept of residual policy learning is to first have a base policy $\mu(s)$ and then use RL to learn a policy $\pi(s)$ that outputs action residual to the first policy. The final action from RPL takes the form of $a = \mu(s) + \pi(s)$. Our instantiation of RPL uses the same deep ResNet-18 BC policies as IBRL and we also allow the RL residual policy to take the output of the BC policy as an additional input to provide it with a useful initial guess, i.e. $a = \mu(s) + \pi(s, \mu(s))$. The BC policy is kept fixed and we optimize the residual policy using the same RL backbone used by all model-free RL methods in this paper. Furthermore, we follow~\cite{rpl} to zero out the last layer of the RL policy in RPL so that the initial actions are close to the BC actions. 

\cref{fig:rpl} shows the performance of RPL alongside IBRL and other baselines in the two Robomimic tasks with image inputs. RPL performs well compared against other baselines but not as well as IBRL. Inspired by the strong performance of RPL, we are interested in understanding if the residual formulation benefits other methods. Therefore, we additionally run IBRL with an RPL-style modification to the input and output of the policy network (IBRL-RPL, dotted line in~\cref{fig:rpl}). We find that IBRL-RPL further improves the sample efficiency over IBRL on the harder Square task and maintains roughly the same performance on the simpler Can task. It is encouraging that IBRL can be combined with existing techniques to achieve even better performance.

\section{Comparison with ROT}
\label{app:rot}

\begin{figure}[h]
\centering
\includegraphics[width=\linewidth]{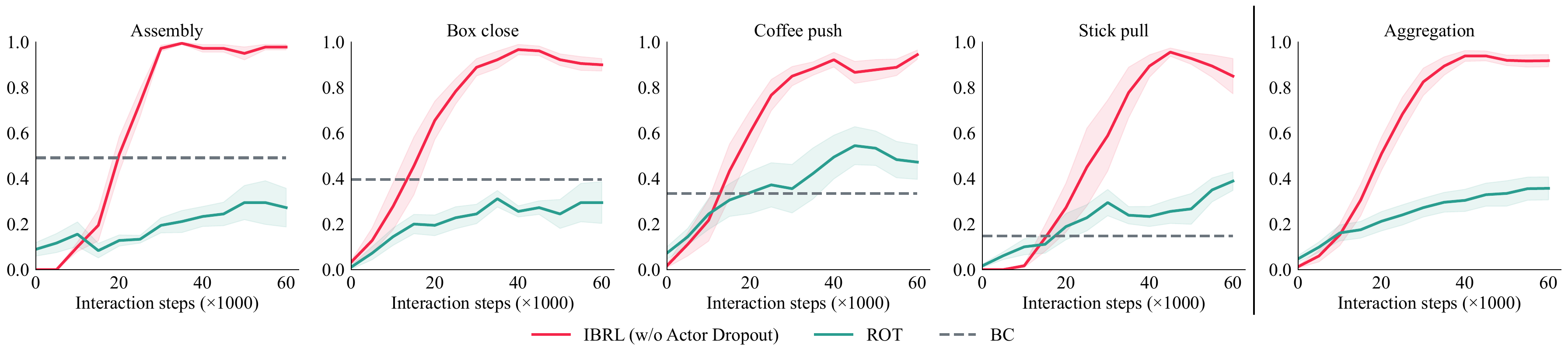}
\caption{Comparison with IBRL and ROT, one of the best performing RL method that does not require environment reward. Note IBRL and ROT have \textbf{different assumptions} because ROT does not use the sparse 0/1 reward from the environment. This comparison is mainly to illustrate the difference in the peak performance under different assumptions (sparse reward v.s. no environment reward at all).}
\label{fig:rot}
\end{figure}

To understand the difference in the peak performance between RL with sparse reward and RL that assumes no access to environment reward at all (such as inverse RL, or online imitation learning), we compare IBRL against ROT~\cite{haldar2022watch-rot}, a powerful online imitation learning method that has shown to outperform a wide range of other inverse RL methods. Our RFT baseline is closely related to ROT. ROT can be seen as RFT without the sparse reward from the environment but instead with a dense trajectory matching reward computed by optimal transport. We emphasize the methods considered in this paper have \textbf{different assumptions} from ROT or inverse RL/online imitation in general as the later family of methods do no assume access of any environment rewards and instead use reward predicted from demonstrations.

~\cref{fig:rot} shows IBRL and ROT on the Meta-World tasks. Unsurprisingly, IBRL performs significantly better than ROT. Note that the Meta-World tasks considered in this paper are harder than the ones considered in the original ROT paper and we also run on significantly smaller sample budgets (60K vs 1M). Additionally, we find that adding OT reward on IBRL or RFT no longer helps but sometimes hurts performance when having access to the ground truth sparse reward as it is challenging to balance the magnitude of the two reward sources.

One takeaway from this experiment is that ground truth reward, even sparse, makes a huge difference in the performance of the RL method. When sparse reward is accurate, IBRL learns efficiently without relying on any dense reward signals. This suggests accurate and robust success prediction as an important research direction for RL on real robots.

\begin{figure}[h]
\centering
\begin{lstlisting}[numbers=none]
VitEncoder(
  (patch_embed): PatchEmbed(
    (embed): Sequential(
      (conv1): Conv2d(3, 128, kernel_size=(8, 8), stride=(4, 4))
      (relu): ReLU()
      (conv2): Conv2d(128, 128, kernel_size=(3, 3), stride=(2, 2))
    )
  )
  (net): Sequential(
    TransformerLayer(
      (layer_norm1): LayerNorm()
      (mha): MultiHeadAttention(
        (qkv_proj): Linear(in_features=128, out_features=384, bias=True)
        (out_proj): Linear(in_features=128, out_features=128, bias=True)
      )
      (layer_norm2): LayerNorm()
      (linear1): Linear(in_features=128, out_features=512, bias=True)
      (linear2): Linear(in_features=512, out_features=128, bias=True)
    )
  )
  (norm): LayerNorm()
)
\end{lstlisting}
\caption{
Architecture of ViT encoder expressed in PyTorch style pseudocode. The shape of the input image is $(3, 96, 96)$ in all experiments. The shape of the output of the ViT encoder is $(121, 128)$, i.e., 121 patches where each patch is a 128-dimensional vector.}
\label{fig:vit-code}
\vspace{-6mm}
\end{figure}

\begin{figure}[h]
\centering
\begin{lstlisting}[numbers=none]
Critic(
  (spatial_embed) SpatialEmbed(
    (weight): Parameter(128, 1024)
    (input_proj): Sequential(
      (0): Linear(in_features=155, out_features=1024, bias=True)
      (1): LayerNorm()
      (2): ReLU(inplace=True)
    )
  )
  (q): Sequential(
    (0): Linear(in_features=1058, out_features=1024, bias=True)
    (1): LayerNorm()
    (2): ReLU(inplace=True)
    (3): Linear(in_features=1024, out_features=1024, bias=True)
    (4): LayerNorm()
    (5): ReLU(inplace=True)
    (6): Linear(in_features=1024, out_features=1, bias=True)
  )
)
\end{lstlisting}
\caption{
Architecture of the critic head expressed in PyTorch style pseudocode. 
We first transpose the output of ViT encoder $(121, 128) \rightarrow (128, 121)$ and then append three most recent proprioception data $(3 \times 8,)$ and the action to evaluate $(7,)$ to each channel. 
Hence the input size of the \texttt{input\_proj} is $(155 = 121 + 3 * 8 + 7)$. 
We apply an element-wise multiplication between the output of \texttt{input\_proj} and \texttt{weight}, and sum over the channel dimension to produce a $1024$-dimensional vector as the output of \texttt{SpatialEmbed}. Finally, we append the action to the output of \texttt{SpatialEmbed} again before feeding it to the Q-MLP.
}
\label{fig:critic-code}
\end{figure}

\newpage
\begin{figure}[h]
\centering
\begin{lstlisting}[numbers=none]
Actor(
  (compress): Sequential(
    (0): Linear(in_features=15488, out_features=128, bias=True)
    (1): LayerNorm()
    (2): Dropout(p=0.5, inplace=False)
    (3): ReLU()
  )
  (policy): Sequential(
    (0): Linear(in_features=155, out_features=1024, bias=True)
    (1): LayerNorm()
    (2): Dropout(p=0.5, inplace=False)
    (3): ReLU()
    (4): Linear(in_features=1024, out_features=1024, bias=True)
    (5): LayerNorm()
    (6): Dropout(p=0.5, inplace=False)
    (7): ReLU()
    (8): Linear(in_features=1024, out_features=7, bias=True)
    (9): Tanh()
  )
)
\end{lstlisting}
\caption{
Architecture of the policy head. It takes the flattened output of the ViT encoder, i.e. $15488 = 121 \times 128$. We append three most recent proprioception data $(3 \times 8,)$ to the output of the \texttt{compress} module before feeding it to the \texttt{policy} module.}
\label{fig:actor-code}
\vspace{-6mm}
\end{figure}

\begin{figure}[h]
\centering
\begin{lstlisting}[numbers=none]
Critic(
  (net): Sequential(
    (0): Linear(in_features=3 * state_dim + action_dim, out_features=1024, bias=True)
    (1): LayerNorm((1024,), eps=1e-05, elementwise_affine=True)
    (2): ReLU()
    (3): Linear(in_features=1024, out_features=1024, bias=True)
    (4): LayerNorm((1024,), eps=1e-05, elementwise_affine=True)
    (5): ReLU()
    (6): Linear(in_features=1024, out_features=1024, bias=True)
    (7): LayerNorm((1024,), eps=1e-05, elementwise_affine=True)
    (8): ReLU()
    (9): Linear(in_features=1024, out_features=1, bias=True)
  )
)

Actor(
  (net): Sequential(
    (0): Linear(in_features=3 * state_dim, out_features=1024, bias=True)
    (1): Dropout(p=0.5, inplace=False)
    (2): ReLU()
    (3): Linear(in_features=1024, out_features=1024, bias=True)
    (4): Dropout(p=0.5, inplace=False)
    (5): ReLU()
    (6): Linear(in_features=1024, out_features=1024, bias=True)
    (7): Dropout(p=0.5, inplace=False)
    (8): ReLU()
    (9): Linear(in_features=1024, out_features=action_dim, bias=True)
    (10): Tanh()
  )
)
\end{lstlisting}
\caption{Architecture of critic and policy network in state-based RL.}
\label{fig:state-code}
\end{figure}

}

\clearpage
\end{document}